\documentclass[11pt]{article}

\usepackage[utf8]{inputenc}
\usepackage[T1]{fontenc}
\usepackage[a4paper,margin=1in]{geometry}
\usepackage{microtype}
\usepackage[table,dvipsnames]{xcolor} 
\usepackage{graphicx}
\usepackage{booktabs}
\usepackage{amsmath,amsfonts}
\usepackage{nicefrac}
\usepackage{caption}
\usepackage{subcaption}
\usepackage{placeins}
\usepackage{natbib}
\usepackage{url}
\usepackage{hyperref}
\usepackage{fontawesome5}
\usepackage{authblk} 

\setcitestyle{authoryear,round}

\hypersetup{
  colorlinks=true,
  linkcolor=MidnightBlue,
  citecolor=MidnightBlue,
  urlcolor=RoyalBlue,
  pdftitle={Process Reward Models for Sentence-Level Verification of LVLM Radiology Reports},
  pdfauthor={Alois Thomas, Maya Varma, Jean-Benoit Delbrouck, Curtis P. Langlotz}
}

\date{} 

\title{Process Reward Models for Sentence-Level Verification of LVLM Radiology Reports}

\author{
  Alois Thomas \quad Maya Varma \quad Jean\mbox{-}Benoit Delbrouck \quad Curtis P.\ Langlotz \\
  AIMI Center, Stanford University, CA, USA \\
  \texttt{\{aloistho, mayavarma, jbdel, langlotz\}@stanford.edu}
}

\begin{document}
\maketitle

\begin{abstract}
Automating radiology report generation with Large Vision-Language Models (LVLMs) holds great potential, yet these models often produce clinically critical hallucinations, posing serious risks. Existing hallucination detection methods frequently lack the necessary sentence-level granularity or robust generalization across different LVLM generators. We introduce a novel approach: a sentence-level Process Reward Model (PRM) adapted for this vision-language task. Our PRM predicts the factual correctness of each generated sentence, conditioned on clinical context and preceding text. When fine-tuned on MIMIC-CXR with weakly-supervised labels, a lightweight 0.5B-parameter PRM outperforms existing verification techniques, demonstrating, for instance, relative improvements of 7.5\% in Matthews Correlation Coefficient and 1.8\% in AUROC over strong white-box baselines on outputs from one LVLM. Unlike methods reliant on internal model states, our PRM demonstrates strong generalization to an unseen LVLM. We further show its practical utility: PRM scores effectively filter low-quality reports, improving F1-CheXbert scores by 4.5\% (when discarding the worst 10\% of reports). Moreover, when guiding a novel weighted best-of-N selection process on the MIMIC-CXR test set, our PRM show relative improvements in clinical metrics of 7.4\% for F1-CheXbert and 0.6\% for BERTScore. These results demonstrate that a lightweight, context-aware PRM provides a model-agnostic safety layer for clinical LVLMs without access to internal activations
\end{abstract}

\section{Introduction}
\label{sec:introduction}

Large Language Models (LLMs) and Vision-Language Models (VLMs or LVLMs) have achieved remarkable success in generating free-text content, including in healthcare \citep{chen2024inside, maira2}. 
Radiology report generation involves translating medical images (like X-rays or CT scans) and patient-specific clinical information into a structured textual summary of findings. Automating this complex task can help alleviate the increasing workload on radiologists and improve patient care.
In radiology, these models offer potential for automating report generation from medical images and patient context, possibly alleviating radiologist workload \citep{paschali2025foundation}. However, a significant challenge remains: the propensity of these models to generate "hallucinations", factually incorrect statements, which pose severe risks in clinical settings where diagnostic accuracy is important \citep{Kim2025.02.28.25323115}.

\begin{figure}[htbp]
  \centering
  \includegraphics[width=0.8\linewidth]{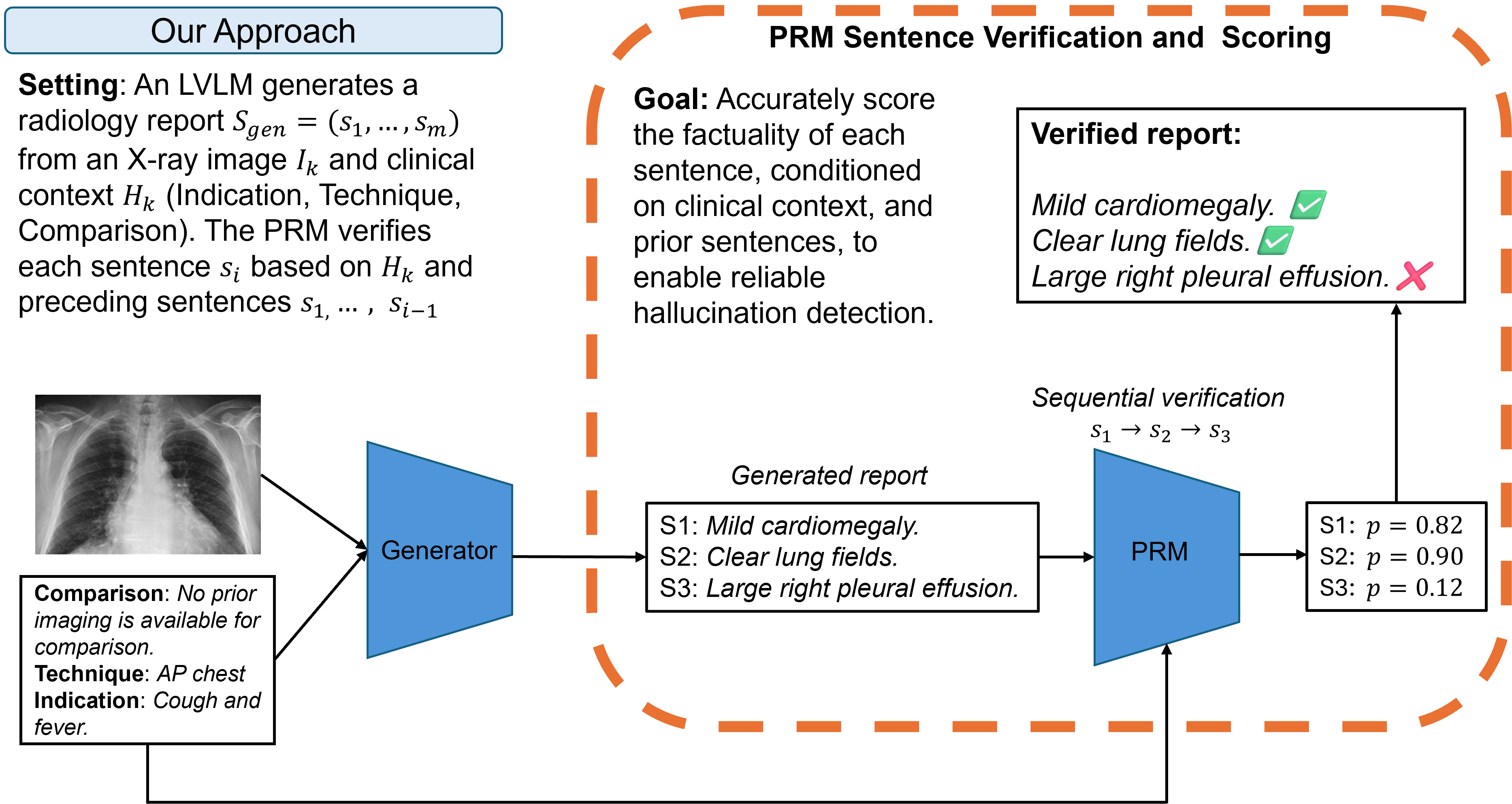}
  \caption{Overview of our proposed sentence-level Process Reward Model (PRM) for verifying radiology reports. The PRM takes clinical context and previously generated/verified sentences as input to predict the correctness of the current sentence, enabling fine-grained hallucination detection.}
  \label{fig:main_figure_intro}
\end{figure}
\FloatBarrier 

Existing hallucination detection methods often operate at the report level or lack robustness across different generator models \citep{chen2024inside}. Black-box methods, relying on sampling consistency or entailment checks, can be computationally expensive \citep{Zhang2024, Manakul2023}. White-box methods, using internal model states like hidden activations or logits, may offer efficiency but often exhibit poor generalization when applied to different LVLMs \citep{Hardy2024, Azaria2023}. 
Neither approach typically provides the sentence-level granularity needed in radiology, where a single incorrect sentence (e.g., misstating the presence of a critical finding like pneumothorax) can have serious clinical consequences, potentially leading to misdiagnosis, delayed or incorrect treatment, and ultimately, adverse patient outcomes. Therefore, identifying and mitigating errors at the sentence level is important for safe clinical deployment.

To address these limitations, we propose adapting Process Reward Models (PRMs), initially developed for evaluating step-by-step reasoning in LLMs \citep{Lightman2023}, to the task of sentence-level factuality verification in radiology reports. 
Instead of assessing mathematical reasoning steps, our modified PRM is a novel adaptation for this multimodal clinical domain, and learns to sequentially evaluate the factual grounding of radiology report sentences. It assigns a correctness probability to each sentence, conditioned on the full clinical context (patient history, imaging technique, comparison studies) and all previously generated and verified sentences in the report. This fine-grained, context-aware verification operates in a black-box manner, requiring only the generated text and clinical prompt, making it broadly applicable.
Our contributions are:
\begin{enumerate}
    \item We propose a novel sentence-level Process Reward Model (PRM) specifically designed for verifying LVLM-generated radiology findings. This PRM operates sequentially, conditioned on clinical context and prior generated text, and functions as a black-box verifier, marking a new application of PRMs to fine-grained, multimodal verification in this critical healthcare domain.
    \item We introduce a scalable training methodology for this PRM using a large corpus ($\sim$200k instances) of weakly-supervised sentence correctness labels derived from an existing domain-specific entailment model (RadNLI).
    \item Through extensive experiments on the MIMIC-CXR dataset \citep{Johnson2019}, we demonstrate that our PRM verifier, leveraging the proposed approach, outperforms strong baselines (including ReXTrust, \citet{Hardy2024}) in sentence-level hallucination detection and exhibits superior transferability to an unseen LVLM.
    \item We further showcase its practical utility by showing how PRM scores can effectively filter low-quality reports via rejection sampling and improve overall report quality via Best-of-N selection, especially using a novel weighted strategy leveraging CheXbert labels.
\end{enumerate}
This work offers a path towards more reliable and clinically safer deployment of LVLMs in radiology.

\section{Approach}
\label{sec:approach}

We formalise sentence-level hallucination detection for radiology reports, and introduce a process reward model (PRM) that performs this verification in-context. We position two strong baselines against it, and describe how the resulting sentence probabilities can lead to practical downstream improvements of radiology LVLMs.

\subsection{Problem statement and notation}
\label{sec:problem_statement}

\textbf{Task.}  For each study \(k\) we observe a chest X-ray image \(I_k\) and structured clinical context \(H_k\) comprising the indication, imaging technique and any comparison studies.  A report generator produces a findings section \(S^{\text{gen}}_{k}=(s^{\text{gen}}_{k,1},\dots,s^{\text{gen}}_{k,m_k})\).  Our goal is to estimate, for every sentence \(s^{\text{gen}}_{k,i}\), the probability that it is factually correct with respect to \(\{I_k,H_k\}\).

\textbf{Supervision.}  
We label sentences automatically with RadNLI \citep{Yuan2021} (which achieves 80.0\% accuracy on its test set, see Appendix~\ref{sec:data_details} Fig.~\ref{fig:nli_comparison}): if any ground-truth sentence semantically entails \(s^{\text{gen}}_{k,i}\) we assign \(y_{k,i}=1\), otherwise \(y_{k,i}=0\).  This yields a noisy but large training corpus that we balance to an approximate 1:1 ratio (details in Appendix~\ref{sec:data_details}).

\subsection{Sentence-level process reward model}
\label{sec:prm_verifier}

\paragraph{Core idea.}
We instantiate the PRM as a sequential binary decoder: given the clinical prompt and the running prefix of previously generated sentences with their predicted labels, the model outputs a token \(y_{k,i}\!\in\!\{0,1\}\) that denotes the factual status of the current sentence.

\paragraph{Input construction.}
During training we provide the model with the interleaved sequence
\[
\bigl[\,H_k,\; s^{\text{gen}}_{k,1},\;\texttt{\textbackslash n},\;y_{k,1},\;\dots,\;s^{\text{gen}}_{k,m_k},\;\texttt{\textbackslash n},\;y_{k,m_k}\bigr],
\]
where \verb|\n| is a hard separator.  At inference the labels are omitted; the verifier must supply them.

\paragraph{Model backbone and scaling.}
We fine-tune Qwen2.5-0.5B-base and Qwen2.5-3B-base \citep{Qwen2024} with the TRL PRM-trainer, thereby evaluating the effect of parameter count on verification quality (training protocol in Section~\ref{sec:model_training}).

\paragraph{Objective.}
Let \(\phi\) denote the PRM parameters and
\(\text{prefix}_{k,i}=[H_k,s^{\text{gen}}_{k,1},\texttt{\textbackslash n},y_{k,1},\dots,s^{\text{gen}}_{k,i},\texttt{\textbackslash n}]\).
Constraining the logits to \{\texttt{‘0’},\texttt{‘1’}\}, we compute
\[
p_{\phi,k,i}\;=\;p_\phi\!\bigl(\texttt{‘1’}\mid\text{prefix}_{k,i}\bigr),
\]
and minimise the binary cross-entropy
\begin{equation}
\label{eq:prm_loss}
\mathcal{L}_\text{PRM}(\phi)=
-\sum_{k}\sum_{i} \Bigl[
      y_{k,i}\,\log p_{\phi,k,i}
      + (1-y_{k,i})\,\log\!\bigl(1-p_{\phi,k,i}\bigr)\Bigr].
\end{equation}
At test time the verifier outputs \(\{p_{\phi,k,i}\}_{i=1}^{m_k}\), a sentence-level plausibility map for the generated report.

\subsection{Baseline verifiers}
\label{sec:baselines}

\textbf{ReXTrust (white-box).}
Following Hardy et al.\ \citep{Hardy2024}, we feed MAIRA-2’s 16-th layer token embeddings for each sentence into a multi-head self-attention block, mean-pool the result, and apply a linear classifier.

\textbf{Grey-box MLP.}
A two-layer perceptron inspired by Liu et al.\ \citep{Liu2024} uses a 13-dimensional feature vector of sentence-level token statistics (length, logit and entropy derived features) and outputs a correctness logit.

Hyper-parameters for both baselines are listed in Appendix~\ref{sec:baseline_training_details}.

\subsection{Downstream applications}
\label{sec:downstream_tasks}

\paragraph{Report rejection.}
Aggregating the sentence-level correctness probabilities produces a scalar report score \(\text{Score}_k = f(\{p_{k,i}\})\), which can be used to filter low-quality reports. Let \(p_{\text{PRM}, k, i}\), \(p_{\text{Rex}, k, i}\), and \(p_{\text{MLP}, k, i}\) denote the predicted correctness probability for sentence $i$ of report $k$ from the PRM, ReXTrust, and MLP verifiers, respectively, and let $m_k$ be the number of sentences in report $k$. For the rejection experiments (Section \ref{sec:results_rejection_short}), we study several aggregation functions $f$ (matching labels like `PRM avg`, `PRM prod` in Fig \ref{fig:chex_rad_reject}):
\begin{itemize} 
    \item \textbf{MinProb} (\textit{min}): The minimum sentence probability across the report ($\text{Score}_k = \min_i p_{\text{PRM}, k, i}$). Applied to PRM, ReXTrust, MLP.
    \item \textbf{AvgProb} (\textit{avg}): The arithmetic mean of sentence probabilities ($\text{Score}_k = \frac{1}{m_k} \sum_i p_{\text{PRM}, k, i}$). Applied to PRM, ReXTrust, MLP.
    \item \textbf{ProdProb} (\textit{prod}): The geometric mean of sentence probabilities ($\text{Score}_k = (\prod_{i=1}^{m_k} p_{\text{PRM}, k, i})^{1/m_k}$). Applied to PRM, ReXTrust, MLP.
    \item \textbf{Entropy} (baseline): The average sentence-level token entropy calculated from the generator model's output probabilities (MAIRA-2). Lower scores indicate higher generator confidence.
\end{itemize}
Reports are ranked by these scores. To evaluate rejection, we discard reports below a tunable percentile threshold (e.g., the worst 10\% based on the chosen score).

\paragraph{Best-of-$N$ selection.}
For each study we sample $N{=}128$ candidate reports from MAIRA-2 (temperature 1.0, top-$p$ 0.9, top-$k$ 50) and rank them by:
(i) generator log-probability;
(ii) PRM-derived scores; or
(iii) a weighted PRM strategy that first groups candidates by CheXbert label vector \citep{Smit2020}, sums PRM scores within each group, and then selects the top-scoring candidate in the best group (full algorithm in Appendix~\ref{sec:bon_details}).
Evaluation relies on BLEU, ROUGE, BERTScore, F1-RadGraph and F1-CheXbert.

Together these components define a pipeline that not only flags hallucinated sentences but also demonstrably improves report quality.

\section{Experimental Setup}
\label{sec:exp_setup}

This section details the data, training protocol, and evaluation methodology used to quantify the effectiveness of our sentence-level PRM verifier relative to strong grey- and white-box baselines.

\subsection{Dataset and preprocessing}
\label{sec:data_preprocessing}

\textbf{Dataset.}
We adopt MIMIC-CXR v2.0.0 \citep{Johnson2019} and adhere to its official train, validation, and test splits.

\textbf{Report generation and sentence labelling.}
For each study we prompt MAIRA-2 to generate a findings section, segment the output into sentences, and determine for every sentence whether it is entailed by at least one ground-truth sentence using the domain-specific RadNLI model \citep{Yuan2021}.  Sentences predicted as entailment receive the label $y{=}1$ (correct); all others are marked $y{=}0$ (hallucinated). Initial labeling revealed class imbalance. We down-sampled the majority class in the training, validation, and test sets to achieve an approximate 1:1 ratio, mitigating potential bias. Full details on the balancing procedure and the final sentence counts are provided in Appendix~\ref{sec:data_details}.

\textbf{Input representations.}
ReXTrust uses hidden states from MAIRA-2’s 16-th layer, while the grey-box baseline relies on a 13-dimensional vector of sentence-level token statistics (length, logit and entropy moments).  The PRM receives a single sequence comprising the clinical prompt followed by each generated sentence, with labels interleaved during training as described in Section \ref{sec:prm_verifier}.  Appendix \ref{sec:data_preparation_details} illustrates the exact formatting.

\textbf{Transferability corpus.}
To assess out-of-distribution generalisation we repeat the above procedure with CheXagent-8B \citep{chen2024vision}, yielding an independent test bed of automatically labelled reports.

\subsection{Model training}
\label{sec:model_training}

All models are trained on the same sentence-balanced splits.
We fine-tune Qwen2.5-0.5B and Qwen2.5-3B with the TRL PRM-trainer for three epochs, using AdamW ($\text{lr}=10^{-5}$) with linear warm-up and decay, an effective batch size of 16 (2 samples per GPU, 8 gradient-accumulation steps), gradient checkpointing, and a maximum sequence length of 1,024 tokens.

For ReXTrust we follow Hardy et al.\ \citep{Hardy2024}: AdamW ($\text{lr}=10^{-4}$), cosine annealing, weight decay 0.01, batch size 128, dropout 0.1, and early stopping after three stagnant validation epochs.
The grey-box MLP (two hidden layers) is optimised with Adam at $\text{lr}=10^{-3}$ for 20 epochs and the same batch size.
Each model minimises binary cross-entropy.
We run the training of PRMs on a single A100 GPU.
Best-of-$N$ experiments run on an 8$\times$H100 GPU cluster. Full hyper-parameters appear in Appendix \ref{sec:training_details_appendix}.

\subsection{Evaluation metrics}
\label{sec:evaluation_metrics}

Sentence-level discrimination is reported via accuracy, macro-averaged F1, Matthews correlation coefficient (MCC), AUROC, and AUPRC.
Report-level quality is assessed with two clinical metrics—F1-CheXbert \citep{Smit2020} and F1-RadGraph \citep{delbrouck2024radgraph}—and three textual metrics (BLEU \citep{papineni2002bleu}, ROUGE \citep{lin2004rouge}, and BERTScore \citep{zhang2019bertscore}).

\subsection{Scope of experiments}
\label{sec:experimental_scope}

We design seven complementary studies:
(i)~baseline comparison on sentence-level metrics;
(ii)~generator transfer to CheXagent-8B;
(iii)~report rejection, where low-scoring reports are discarded;
(iv)~Best-of-128 sampling, including our weighted selection strategy;
(v)~context ablation to isolate which prompt fields are indispensable; and
(vi)~model-scaling analysis contrasting 0.5B and 3B PRMs.

\section{Results}
\label{sec:results_short}
\FloatBarrier

We present results evaluating the PRM verifier against baselines on verification, transferability, and downstream tasks.

\subsection{Sentence-level verification performance}
\label{sec:results_sentence_level_short}
\FloatBarrier

We compared our PRM verifiers (Qwen2.5-0.5B, Qwen2.5-3B) against the grey-box MLP \citep{Liu2024} and the white-box ReXTrust \citep{Hardy2024} on the MIMIC-CXR test set. Table \ref{tab:main_train_results} shows performance with 95\% CIs (1000 bootstrap resamples).

\begin{table}[htbp]
\centering
\caption{Sentence-level verification performance on the MIMIC-CXR test set. Metrics computed using 1,000 bootstrap resamples with 95\% confidence intervals.}
\label{tab:main_train_results}
\resizebox{\linewidth}{!}{
\begin{tabular}{@{}lccccc@{}}
\toprule
\textbf{Model} & \textbf{Accuracy} & \textbf{F1 Macro} & \textbf{MCC} & \textbf{AUROC} & \textbf{AUPRC} \\
\midrule
Grey-box       & \shortstack{0.652 \\ (0.628, 0.677)} & \shortstack{0.646 \\ (0.622, 0.669)} & \shortstack{0.300 \\ (0.253, 0.345)} & \shortstack{0.701 \\ (0.674, 0.727)} & \shortstack{0.684 \\ (0.647, 0.721)} \\
ReXTrust       & \shortstack{0.735 \\ (0.712, 0.755)} & \shortstack{0.733 \\ (0.710, 0.754)} & \shortstack{0.467 \\ (0.422, 0.510)} & \shortstack{0.819 \\ (0.798, 0.839)} & \shortstack{0.798 \\ (0.768, 0.828)} \\
\rowcolor{gray!20} Qwen2.5-0.5B-PRM & \shortstack{\textbf{0.752} \\ (0.728, 0.773)} & \shortstack{\textbf{0.751} \\ (0.727, 0.771)} & \shortstack{\textbf{0.502} \\ (0.455, 0.544)} & \shortstack{0.834 \\ (0.815, 0.853)} & \shortstack{0.832 \\ (0.806, 0.856)} \\
\rowcolor{gray!20} Qwen2.5-3B-PRM   & \shortstack{0.746 \\ (0.723, 0.771)} & \shortstack{0.744 \\ (0.722, 0.769)} & \shortstack{0.490 \\ (0.444, 0.539)} & \shortstack{\textbf{0.841} \\ (0.821, 0.861)} & \shortstack{\textbf{0.835} \\ (0.811, 0.861)} \\
\bottomrule
\end{tabular}
}
\end{table}

The PRM models consistently outperform both baselines. Qwen2.5-0.5B-PRM achieves the highest Accuracy, F1 Macro, and MCC (+3.5 MCC points over ReXTrust). Qwen2.5-3B-PRM yields the best AUROC (+2.2 points over ReXTrust) and AUPRC, indicating superior discrimination despite slightly lower accuracy/F1 than the 0.5B PRM. This shows the effectiveness of the PRM approach, potentially leveraging contextual understanding better than hidden-state or grey-box derived methods. Qualitative examples are in Appendix \ref{sec:appendix_qualitative}.

\paragraph{Keyword-specific performance.} We assessed F1-micro scores for sentences containing specific keywords (Table \ref{tab:keyword_f1}). PRM variants generally outperform ReXTrust on most keywords (e.g., "pleural effusion", "consolidation", "pneumothorax"). Qwen2.5-0.5B-PRM leads on "edema", "atelectasis", "left", while Qwen2.5-3B-PRM excels on "tube", "right". This shows PRM superiority extends to clinically relevant terms.

\begin{table}[htbp]
\centering
\caption{F1-micro scores on sentences containing selected keywords (test set), 95\% CIs.}
\label{tab:keyword_f1}
\resizebox{\textwidth}{!}{
\begin{tabular}{@{}lcccc@{}}
\toprule
\textbf{Keyword} & \textbf{Count} & \textbf{Qwen2.5-0.5B-PRM F1} & \textbf{Qwen2.5-3B-PRM F1} & \textbf{ReXTrust F1} \\
\midrule
``pleural effusion'' & 218 & \textbf{0.789} (0.734, 0.839) & 0.780 (0.725, 0.830) & 0.775 (0.716, 0.830) \\
``pneumothorax''     & 215 & 0.865 (0.819, 0.907) & \textbf{0.874} (0.833, 0.916) & 0.870 (0.823, 0.907) \\
``consolidation''    & 125 & \textbf{0.664} (0.576, 0.752) & 0.632 (0.544, 0.720) & 0.648 (0.568, 0.728) \\
``pneumonia''        & 5   & \textbf{0.800} (0.400, 0.800) & \textbf{0.800} (0.400, 0.800) & \textbf{0.800} (0.400, 0.800) \\
``edema''            & 51  & \textbf{0.686} (0.549, 0.804) & 0.667 (0.529, 0.785) & 0.608 (0.471, 0.745) \\
``atelectasis''      & 26  & \textbf{0.962} (0.846, 1.000) & 0.923 (0.808, 0.962) & 0.923 (0.808, 0.962) \\
``tube''             & 42  & 0.690 (0.571, 0.833) & \textbf{0.786} (0.667, 0.905) & 0.738 (0.595, 0.857) \\
``right''            & 134 & 0.672 (0.597, 0.746) & \textbf{0.739} (0.664, 0.806) & \textbf{0.739} (0.664, 0.813) \\
``left''             & 93  & \textbf{0.774} (0.688, 0.860) & 0.699 (0.613, 0.785) & 0.710 (0.624, 0.796) \\
\bottomrule
\end{tabular}
}
\end{table}

\subsection{Transferability and generalization}
\label{sec:results_transferability_short} 
\FloatBarrier

We tested models trained on MAIRA-2 reports on reports from CheXagent-8B \citep{chen2024vision} (Table \ref{tab:chexagent_metrics}).

\begin{table}[htbp]
\centering
\caption{Transferability performance on CheXagent-8B reports (95\% CIs).}
\label{tab:chexagent_metrics}
\resizebox{0.9\linewidth}{!}{
\begin{tabular}{@{}lccc@{}}
\toprule
\textbf{Metric} & \textbf{Qwen2.5-0.5B-PRM} & \textbf{Qwen2.5-3B-PRM} & \textbf{ReXTrust} \\
\midrule
Accuracy  & 0.700 (0.673, 0.724)  & 0.699 (0.672, 0.724)  & \textbf{0.805} (0.783, 0.825) \\
F1 Macro  & 0.599 (0.567, 0.629)  & \textbf{0.627} (0.597, 0.654)  & 0.446 (0.439, 0.452) \\
MCC       & 0.220 (0.156, 0.279)  & \textbf{0.306} (0.250, 0.358)  & 0.000 (0.000, 0.000)$^\dagger$ \\
AUROC     & 0.729 (0.697, 0.760)  & \textbf{0.754} (0.722, 0.781)  & 0.411 (0.372, 0.450) \\
AUPRC     & 0.368 (0.307, 0.425)  & \textbf{0.379} (0.322, 0.437)  & 0.169 (0.144, 0.196) \\
\bottomrule
\end{tabular}
} 
\end{table}
{\footnotesize $^\dagger$ An MCC of 0.000 occurs when a binary classifier predicts only a single class for all samples. Here, ReXTrust failed to identify any hallucinated sentences (predicted all as non-hallucinated), resulting in this degenerate MCC value.}

ReXTrust achieves high overall accuracy but has weaker performance on balanced metrics (MCC=0, AUROC<0.5), likely due to majority class prediction, indicating poor generalizability. In contrast, both PRM models maintain strong performance on balanced metrics (e.g., Qwen2.5-3B-PRM MCC=0.306, AUROC=0.754), demonstrating superior transferability. This suggests PRMs learn more generalizable factuality patterns than white-box methods reliant on specific generator internals.

\subsection{Rejecting low-quality reports}
\label{sec:results_rejection_short}

\begin{figure}[htbp]
  \centering
  \begin{subfigure}{0.48\linewidth}
    \centering
    \includegraphics[width=\linewidth]{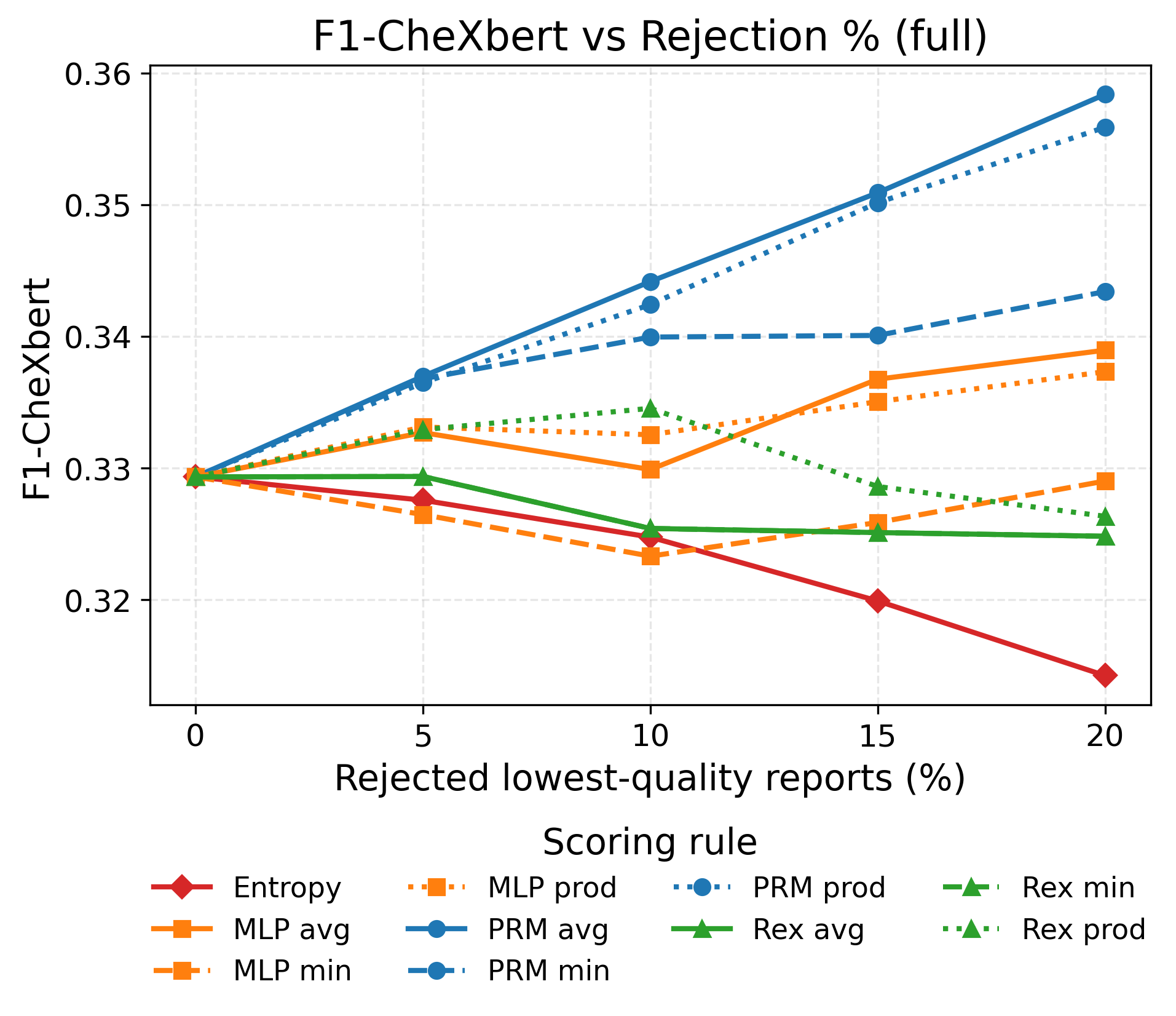}
    \caption{F1-CheXbert (full test set)}
    \label{fig:chex_reject}
  \end{subfigure}\hfill
  \begin{subfigure}{0.48\linewidth}
    \centering
    \includegraphics[width=\linewidth]{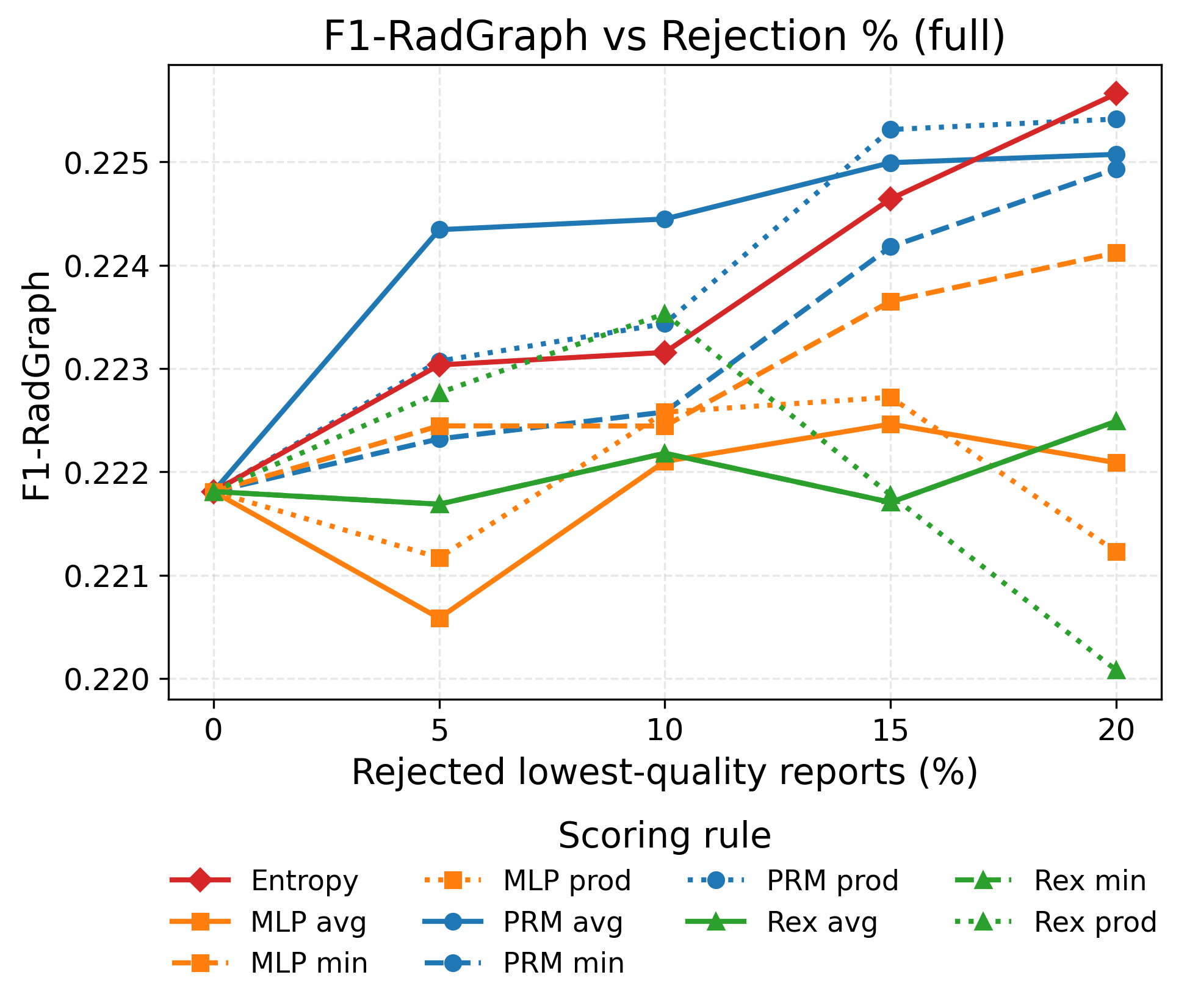}
    \caption{F1-RadGraph (full test set)}
    \label{fig:rad_reject}
  \end{subfigure}
  \caption{Impact of rejecting lowest-scoring reports (x-axis) on clinical factuality metrics (y-axis) of remaining reports. PRM-\textit{avg} and PRM-\textit{prod} show steepest improvements, indicating effective quality filtering. Entropy also shows good filtering quality for F1-RadGraph scores, sometimes outperforming PRM methods at high rejection thresholds.}
  \label{fig:chex_rad_reject}
\end{figure}

We used aggregated sentence probabilities (Min Prob, Avg Prob, Prod Prob) from PRM and baselines to score and reject low-quality reports. Figure \ref{fig:chex_rad_reject} shows the impact on F1-CheXbert and F1-RadGraph as rejection rate increases.

PRM-based scoring, particularly Avg Prob and Prod Prob, generally improves report quality as more low-scoring reports are rejected (steeper curves in Fig~\ref{fig:chex_rad_reject}). For instance, discarding the worst-scoring 10\,\% of reports using PRM-Avg raises F1-CheXbert from 0.329 (baseline F1-CheXbert 0.32934) to 0.344 (0.34417), a relative increase of $+4.5$\,\%. Full results across all rejection thresholds (0/5/10/15/20\,\%) for both F1-CheXbert and F1-RadGraph, along with scores for all aggregation methods, are detailed in App.~\ref{app:all_metrics}, Tab.~\ref{tab:rejection_full_metrics_appendix}. This confirms PRM scores can effectively identify hallucination-prone reports. The baseline Entropy scoring method also demonstrates strong performance for F1-RadGraph, sometimes exceeding PRM-based aggregation at higher rejection thresholds (Fig.~\ref{fig:chex_rad_reject}b and Tab.~\ref{tab:rejection_full_metrics_appendix}).
\FloatBarrier  
\subsection{Best-of-N sampling}
\label{sec:results_bon_short}
\FloatBarrier

\begin{figure}[htbp]
  \centering
  \begin{subfigure}{0.48\linewidth}
    \centering
    \includegraphics[width=\linewidth]{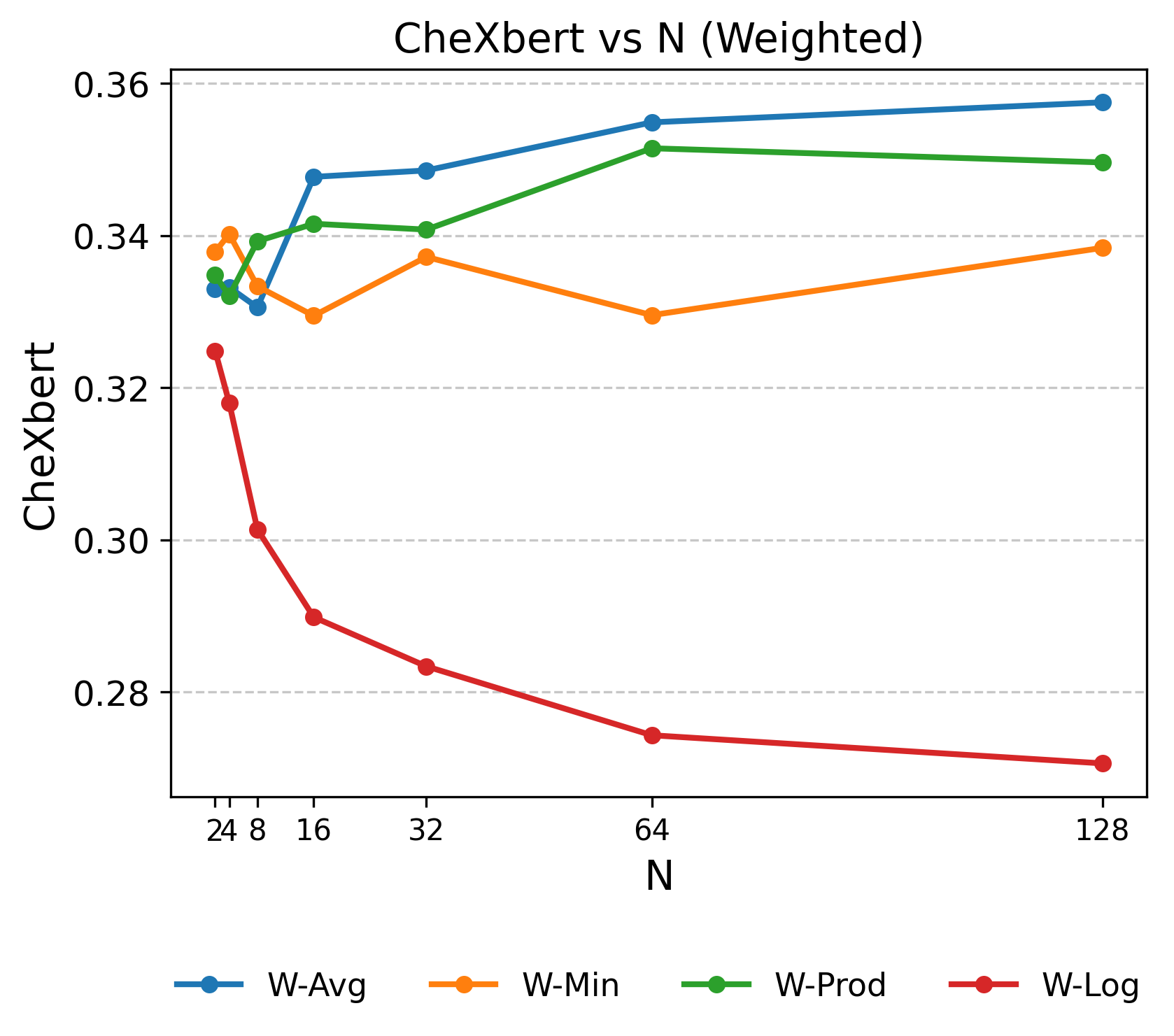}
    \caption{F1-CheXbert (Weighted BoN)}
    \label{fig:bon_chex_weighted}
  \end{subfigure}\hfill
  \begin{subfigure}{0.48\linewidth}
    \centering
    \includegraphics[width=\linewidth]{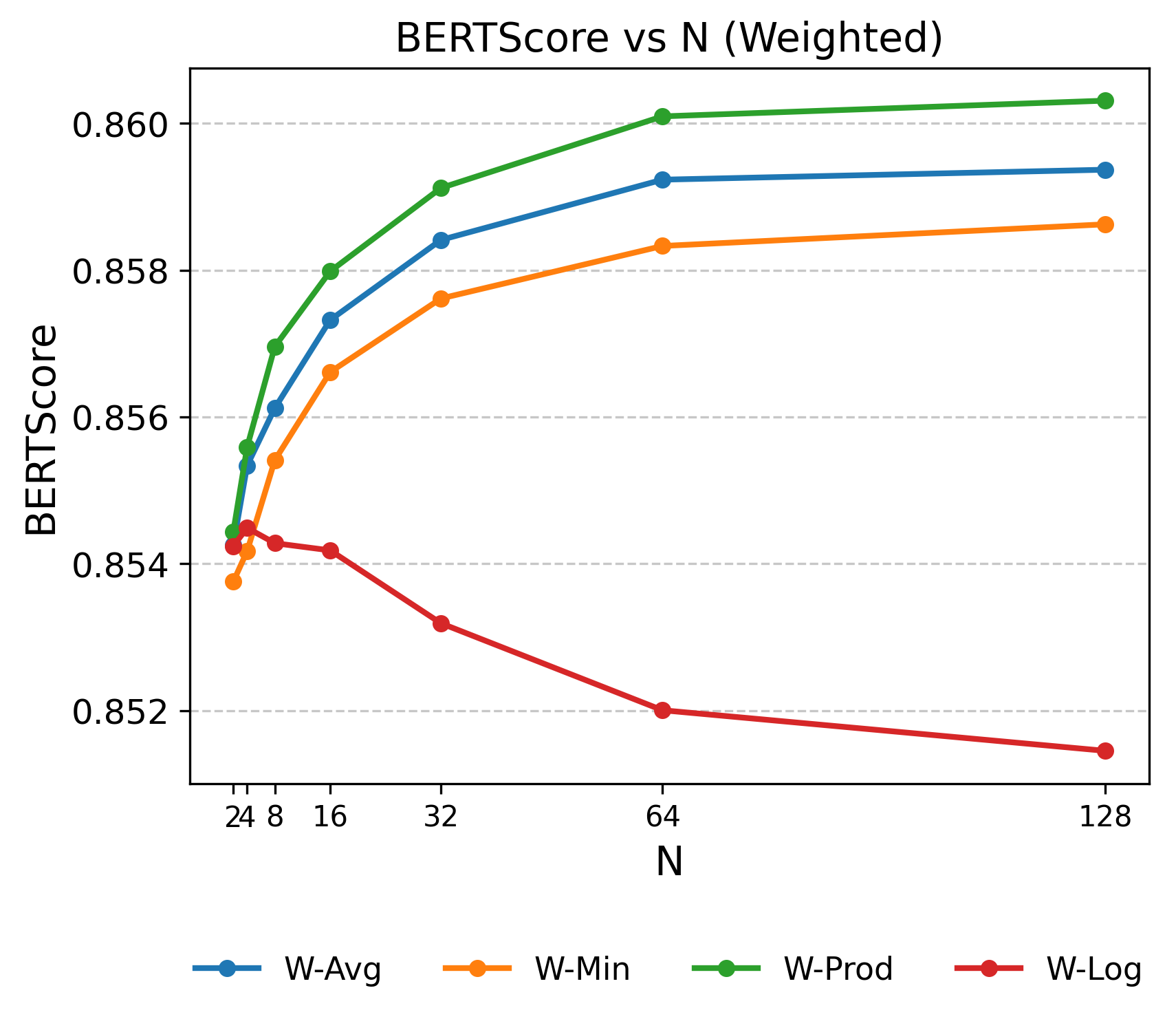}
    \caption{BERTScore (Weighted BoN)}
    \label{fig:bon_bert_weighted}
  \end{subfigure}
  \caption{Performance of Weighted BoN strategies vs. N (temp=1.0). W-Avg and W-Min show strong improvement. Scoring strategies W-Avg, W-Min, W-Prod, and W-Log refer to weighted BoN using base scores from AvgProb, MinProb, ProdProb, and LogProb, respectively.}
  \label{fig:bon_weighted_key}
\end{figure}

We used PRM scores for Best-of-N (BoN) selection from $N=128$ candidates (MAIRA-2, temp=1.0). Figure \ref{fig:bon_weighted_key} shows key results for weighted strategies.

Increasing $N$ generally improves quality. Weighted BoN strategies consistently outperform non-weighted ones (details in Appendix~\ref{app:bon_plots}). W-Avg (Weighted AvgProb) and W-Min (Weighted MinProb) yield strong results across clinical and textual metrics (Fig.~\ref{fig:bon_weighted_key} and App.~\ref{app:bon_plots}, Fig.~\ref{fig:bon_weighted_all}). At N=128, W-Avg achieves an F1-CheXbert of 0.357 and a BERTScore of 0.859. This improves over non-weighted selection (e.g., the non-weighted AvgProb strategy yields F1-CheXbert of 0.308 and BERTScore of 0.858 at N=128, see App.~\ref{app:bon_plots}, Fig.~\ref{fig:bon_nonweighted_all}). Compared to non-weighted AvgProb, W-Avg provides a substantial +5.0 point improvement in F1-CheXbert (0.357 vs 0.308) and a +0.001 point improvement in BERTScore (0.8594 vs 0.8584) at N=128. This demonstrates PRM scores, especially in a weighted BoN framework considering clinical finding profiles (CheXbert labels), effectively select higher-quality, factually accurate reports. Further details about this experiment is available in Appendix~\ref{app:bon_plots}.

\subsection{Ablation studies}
\label{sec:results_ablation_short}
\FloatBarrier

We removed individual context sections (“\textsc{Indication:}”, “\textsc{Technique:}”, “\textsc{Comparison:}”) from the PRM input at inference time (Table~\ref{tab:ablation_results_final}).

Removing \textsc{Technique} produces the largest degradation for both verifiers (e.g.\ Qwen2.5-0.5B: AUROC decreases of 0.007; Qwen2.5-3B: AUROC decreases of 0.018). Eliminating \textsc{Indication} has a comparable negative impact.
The effect of \textsc{Comparison} is size-dependent: for the 0.5 B model its removal is mildly harmful ($\Delta$MCC = –0.012), whereas the 3B model actually improves across every metric (e.g.\ Accuracy + 0.004, MCC + 0.007, AUPRC + 0.001). A plausible explanation is that few training reports contain a well-formed comparison section, and many of those refer to prior studies that are unavailable at inference. For the larger network,whose capacity lets it model subtler correlations, this inconsistent information may act as noise, so discarding it sharpens the verification signal.

\begin{table}[htbp]
\centering
\caption{Ablation results (Balanced test set, 95 \% CIs). Bold indicates the highest value within each model for every metric.}
\label{tab:ablation_results_final}
\resizebox{\linewidth}{!}{%
\begin{tabular}{@{}llccccc@{}}
\toprule
\textbf{Model} & \textbf{Variant} & \textbf{Accuracy} & \textbf{F1 Macro} & \textbf{MCC} & \textbf{AUROC} & \textbf{AUPRC} \\
\midrule
Qwen2.5-0.5B & Original          & \shortstack{\textbf{0.752} \\ (0.728, 0.773)} & \shortstack{\textbf{0.751} \\ (0.727, 0.771)} & \shortstack{\textbf{0.502} \\ (0.455, 0.544)} & \shortstack{\textbf{0.834} \\ (0.815, 0.853)} & \shortstack{\textbf{0.832} \\ (0.806, 0.856)} \\
             & Ablate INDICATION & \shortstack{0.743 \\ (0.720, 0.767)} & \shortstack{0.741 \\ (0.718, 0.765)} & \shortstack{0.483 \\ (0.438, 0.531)} & \shortstack{0.823 \\ (0.802, 0.844)} & \shortstack{0.813 \\ (0.783, 0.841)} \\
             & Ablate TECHNIQUE  & \shortstack{0.737 \\ (0.715, 0.760)} & \shortstack{0.734 \\ (0.712, 0.757)} & \shortstack{0.473 \\ (0.428, 0.517)} & \shortstack{0.827 \\ (0.806, 0.846)} & \shortstack{0.814 \\ (0.783, 0.842)} \\
             & Ablate COMPARISON & \shortstack{0.746 \\ (0.724, 0.768)} & \shortstack{0.745 \\ (0.722, 0.767)} & \shortstack{0.490 \\ (0.446, 0.535)} & \shortstack{0.833 \\ (0.812, 0.852)} & \shortstack{0.827 \\ (0.801, 0.851)} \\
\midrule
Qwen2.5-3B   & Original          & \shortstack{0.746 \\ (0.723, 0.771)} & \shortstack{0.744 \\ (0.722, 0.769)} & \shortstack{0.490 \\ (0.444, 0.539)} & \shortstack{\textbf{0.841} \\ (0.821, 0.861)} & \shortstack{0.835 \\ (0.811, 0.861)} \\
             & Ablate INDICATION & \shortstack{0.743 \\ (0.721, 0.765)} & \shortstack{0.739 \\ (0.717, 0.762)} & \shortstack{0.484 \\ (0.440, 0.530)} & \shortstack{0.825 \\ (0.805, 0.846)} & \shortstack{0.810 \\ (0.779, 0.841)} \\
             & Ablate TECHNIQUE  & \shortstack{0.739 \\ (0.716, 0.760)} & \shortstack{0.737 \\ (0.714, 0.759)} & \shortstack{0.476 \\ (0.429, 0.520)} & \shortstack{0.823 \\ (0.802, 0.843)} & \shortstack{0.806 \\ (0.777, 0.835)} \\
             & Ablate COMPARISON & \shortstack{\textbf{0.750} \\ (0.728, 0.771)} & \shortstack{\textbf{0.748} \\ (0.726, 0.770)} & \shortstack{\textbf{0.497} \\ (0.454, 0.541)} & \shortstack{\textbf{0.841} \\ (0.821, 0.861)} & \shortstack{\textbf{0.836} \\ (0.809, 0.861)} \\
\bottomrule
\end{tabular}%
}
\end{table}

Removing \textsc{Technique} causes the largest performance drop across models and metrics (e.g., Qwen2.5-3B AUROC drops from 0.841 to 0.823). Removing  \textsc{Indication} also noticeably degrades performance. Ablating  \textsc{Comparison} has minimal impact, suggesting the PRM primarily verifies consistency with the current study context and generator text, and less so with prior studies.

\section{Discussion}
\label{sec:discussion}

\paragraph{Verification accuracy.} On the MIMIC–CXR test set our smallest verifier (Qwen2.5‑0.5B) achieved MCC~$0.502$ (95\%~CI:~0.455–0.544) and AUROC~$0.834$ (95\%~CI:~0.815–0.853), exceeding the strongest baseline (ReXTrust: MCC~$0.467$, AUROC~$0.819$) by $+0.035$ and $+0.015$, respectively (Table~\ref{tab:main_train_results}). Interestingly, while the 0.5B model led on threshold-dependent metrics like MCC, the larger 3B model attained the best AUROC/AUPRC. This could indicate that the smaller model better fits the specific classification threshold implied by the binary RadNLI labels, while the larger model learns a more robust underlying ranking of sentence correctness, reflected in the threshold-agnostic metrics.

\paragraph{Generalisation to unseen generators.} When applied to CheXagent‑8B with reports out‑of‑distribution relative to training, the PRM retained balanced discrimination (MCC~$0.306$, AUROC~$0.754$). In contrast, ReXTrust degraded to chance level (MCC~$0.000$). The drop is consistent with the dependency of white‑box methods on generator‑specific latent representations, which may not preserve class‑separable structure after a distribution shift.

\paragraph{Impact on downstream report quality.} Rejecting the lowest‑scoring 10\% of MAIRA‑2 outputs according to the PRM’s mean probability increases F1‑CheXbert from $0.329$ to $0.344$ (+4.5\%, relative). Within a Best‑of‑128 sampling protocol, a weighted selection guided by PRM scores yields F1‑CheXbert $0.357$, a substantial relative improvement over standard log‑probability ranking and non-weighted PRM selection, indicating tangible clinical benefit (Section~\ref{sec:results_bon_short}, Appendix~\ref{app:all_metrics}, Appendix~\ref{app:bon_plots}).

\paragraph{Context ablations.}
\textsc{Technique} and \textsc{Indication} are indispensable as removing either lowers AUROC by up to 0.018 and MCC by up to 0.029 (Table~\ref{tab:ablation_results_final}). In contrast, \textsc{Comparison} has asymmetric effects: for the 0.5B verifier its removal mildly degrades AUROC ($\Delta$AUROC $=-0.001$), whereas the 3B model actually improves across metrics (e.g.\ $\Delta$MCC $=+0.007$). We hypothesise that the comparison section contributes noise because (i) only about 50.6\% of training reports include a substantive comparison (with similar rates in validation (50.0\%) and test (49.1\%)), and (ii) these sentences often reference external priors non related to the current study. The larger model’s greater capacity thus allows it to exploit the cleaner context once this noisy section is removed.

\paragraph{Limitations.}
Our work has several limitations.
First, the sentence correctness labels used for training were generated automatically by RadNLI and are therefore noisy. Using cleaner supervision, such as human annotations, could improve verifier performance.
Second, our evaluation focused on the MIMIC-CXR dataset and reports from two specific generators. Testing on additional datasets like CheXpert or PadChest and with other generators is necessary to establish broader external validity.
Lastly, to annotate the dataset we considered only entailment as a measure of correctness, meaning that correct statements generated by the LVLM that are not present in the ground truth are ultimately labeled as hallucinations. Leveraging LLMs to annotate or human evaluators could mitigate this issue.

\paragraph{Future work.}

Future research could investigate how to calibrate the PRM's output probabilities, which could lead to more reliable thresholds for decision-making.
Another avenue is integrating the PRM as a reward signal within a reinforcement learning framework to directly improve the factuality of generated reports during decoding.
Furthermore, the verification approach could be extended to handle the impression section of radiology reports, which often involves synthesizing findings and clinical reasoning.
Finally, incorporating the images of the studies directly as context for the PRM is another lead to further improve performance.

\section{Conclusion}
\label{sec:conclusion}

Our experiments demonstrate that a lightweight, context‑aware PRM provides a robust detector of sentence‑level hallucinations in LVLM‑generated radiology reports, outperforming grey‑ and white‑box baselines and maintaining discrimination under generator shift. The verifier not only improves clinical metrics through rejection and sampling strategies but also offers a model‑agnostic safety layer that can be attached to any report‑generation pipeline.

\section*{Code availability}
Code and model weights can be found at \href{https://github.com/alothomas/radiology_prm_verifier}{alothomas/radiology\_prm\_verifier}

\bibliography{references} 

\appendix

\section{Related Work}
\label{app:related_work_detailed}
\FloatBarrier 

Hallucination detection in large language models (LLMs) and Large Vision-language models (LVLMs) has attracted considerable attention given its importance in safety-critical applications. Owing to the architectural similarities between LLMs and LVLMs, many techniques are transferable across these domains (\citet{li-etal-2024-dawn}). In the following, we review approaches from two complementary perspectives: (i) \textbf{black-box methods}, which require only model inputs and outputs, and (ii) \textbf{white and grey box methods}, which leverage internal model representations and sampling information.

\subsection{Black-Box Approaches}
\FloatBarrier 

Black-box methods are appealing due to their model-agnostic nature but typically incur higher computational costs.

\paragraph{Sampling-Based Methods.}
Sampling-based techniques generate multiple outputs for a given input to assess consistency. For example, \citet{Zhang2024} introduced RadFlag in the radiology domain, where multiple sampled reports are compared via an entailment model to derive a consistency score. Similarly, entropy-based methods \citet{Farquhar2024} compute semantic entropy over sampled outputs to quantify uncertainty, though at the expense of increased computational demand.

\paragraph{LLMs as Self-Judges.}
An alternative strategy uses LLMs to evaluate their own outputs. Self-critique methods such as SelfCheckGPT \citet{Manakul2023} prompt the LLM to assess consistency between an initial response and subsequent samples, with metrics like BERTScore serving as proxies for factual correctness.

\paragraph{Reward Models.}
Process Reward Models (PRMs) and Outcome Reward Models (ORMs) provide a distinct verification approach by leveraging pre-trained LLMs. Originally proposed for mathematical reasoning \citet{uesato2022solving}, PRMs assess correctness at each intermediate step and have demonstrated superior performance over ORMs in certain tasks \citet{Lightman2023}. Recent work has further explored PRMs for online policy improvement \citet{setlur2024rewarding} and clinical note verification \citet{wang2024process}.

While PRMs have seen applications in general multimodal reasoning and other clinical text scenarios \citep{wang2024process}, their adaptation for fine-grained, sentence-by-sentence factuality assessment of LVLM-generated radiology reports, conditioned on structured clinical context and evolving report text, represents a novel application area which we explore.

\subsection{White-Box and Grey-Box Approaches}
\FloatBarrier

White-box and grey-box methods tap into internal model dynamics to detect hallucinations more directly.

\paragraph{Manipulating Token Probabilities.}
Grey-box approaches modify token log probabilities during generation to steer outputs towards factuality. For instance, \citet{lee2023factuality} showed that greedy decoding can reduce hallucination risk, while \citet{Nguyen2024} proposed min-p sampling, which adjusts the token sampling threshold based on model confidence.

\paragraph{Hidden State Methods.}
White-box techniques leverage latent information in hidden states or logits. Classifiers trained on hidden layer activations have been shown to predict factuality effectively \citet{Azaria2023, ji2024llm}. In radiology, \citet{Hardy2024} utilized token embeddings from intermediate layers (ReXTrust) to detect hallucinated findings, and \citet{chen2024inside} proposed EigenScore, a geometrically motivated measure that computes eigenvalues of the covariance matrix of sentence embeddings. These methods often lack cross-model generalizability, a limitation our PRM approach aims to overcome.

\FloatBarrier

\section{Reproducibility Statement}
\label{sec:appendix_reproducibility}
All datasets, pretrained weights, training scripts, and configuration files necessary to reproduce every table and figure will be released after submission. Detailed instructions cover data acquisition (MIMIC‑CXR v2.0.0 via PhysioNet), label generation, hyper‑parameters, and hardware requirements (8×~NVIDIA H100 and A100).

\section{Ethical Considerations Statement}
\label{sec:appendix_ethics}
MIMIC‑CXR consists of de‑identified images and reports collected under IRB waiver; we adhered to its data‑use agreement and did not attempt re‑identification. The proposed verifier reduces, but does not eliminate, the risk of hallucination. Model‑ and data‑induced biases (e.g., demographic performance gaps) remain unmeasured and should be carefully assessed before any clinical integration.
 
\section{Dataset creation and labeling details}
\label{sec:data_details}
\FloatBarrier

\paragraph{Dataset source}
We used MIMIC-CXR v2.0.0 \citep{Johnson2019} from PhysioNet, adhering to official train/validation/test splits.

\paragraph{Report generation}
Findings sections were generated using MAIRA-2 \citep{maira2} (default settings) and CheXagent-8B \citep{chen2024vision} for transferability tests, using the image and clinical context (indication, technique, comparison) from MIMIC-CXR.

\paragraph{Sentence labeling procedure}
Generated findings were segmented into sentences. Each generated sentence $s_{k,i}^{\text{gen}}$ was compared against all ground truth sentences $s_{k,j}^{\text{gt}}$ using RadNLI \citep{Yuan2021}. If RadNLI predicted `entailment` for any pair $(s_{k,i}^{\text{gen}}, s_{k,j}^{\text{gt}})$, the label was $y_{k,i} = 1$ (correct); otherwise, $y_{k,i} = 0$ (hallucinated). We found RadNLI effective for this domain-specific task compared to general models like GPT-4o (see Figure \ref{fig:nli_comparison}).

\begin{figure}[htbp]
    \centering
    \begin{subfigure}[b]{0.9\textwidth} 
        \centering
        \begin{tabular}{lcc}
            \toprule
            \textbf{Metric} & \textbf{RadNLI model} & \textbf{GPT-4o} \\
            \midrule
            Accuracy  & \textbf{0.8000}  & 0.7750 \\
            F1 macro  & 0.7582 & \textbf{0.7801} \\
            \bottomrule
        \end{tabular}
        \caption{Performance on RadNLI test dataset.}
        \label{subfig:NLI_perf}
    \end{subfigure}

    \begin{subfigure}[b]{0.9\textwidth} 
        \centering
        \fbox{
            \begin{minipage}{0.9\textwidth} 
                You are a highly accurate Natural Language Inference (NLI) classifier and an expert in radiology.\\
                Given two sentences, determine their relationship as \textbf{‘entailment’, ‘neutral’, or ‘contradiction’}.\\
                Respond with only one of these three labels based on the relationship between the sentences.\\[8pt] 
                \texttt{Sentence 1: \{sentence1\}}\\
                \texttt{Sentence 2: \{sentence2\}}\\[5pt] 
                \texttt{Label:}
            \end{minipage}
        }
        \caption{Prompt used for GPT-4o on RadNLI dataset.}
        \label{subfig:GPT4o_prompt}
    \end{subfigure}

    \caption{Comparison of RadNLI \citep{Yuan2021} and GPT-4o for entailment labeling, used for generating weak labels. (a) Performance metrics evaluated on the RadNLI test dataset. (b) Prompt used for GPT-4o evaluation on the RadNLI test dataset.}
    \label{fig:nli_comparison}
\end{figure}

\paragraph{Dataset balancing}
Initial labeling of sentences generated by MAIRA-2 using RadNLI  gives an unbalanced result across the official MIMIC-CXR splits. To mitigate potential bias during training and evaluation, we balanced the training, validation, and test sets. This was achieved by downsampling the majority class (sentences labeled as correct, $y=1$) to achieve roughly a 1:1 ratio with the minority class (hallucinated sentences, $y=0$). The final sentence counts after balancing for each split are presented in Table~\ref{tab:balanced_counts}.

\begin{table}[htbp]
  \centering
  \caption{Final sentence counts per split (train, validation, test) for MAIRA-2 \citep{maira2} generated reports on MIMIC-CXR \citep{Johnson2019} after downsampling the majority class to achieve an approximate 1:1 ratio of correct ($y=1$) vs. hallucinated ($y=0$) sentences, based on RadNLI labels.}
  \label{tab:balanced_counts}
  \begin{tabular}{lrr}
    \toprule
    Split & Correct ($y=1$) & Hallucinated ($y=0$) \\
    \midrule
    Train & 96,255 & 106,950 \\
    Validation & 780 & 865 \\
    Test & 694 & 771 \\
    \bottomrule
  \end{tabular}
\end{table}

\section{Input data preparation details}
\label{sec:data_preparation_details}
\FloatBarrier

\paragraph{ReXTrust baseline input}
For sentence $s_{k,i}^{\text{gen}}$, final hidden states for each token were extracted from MAIRA-2's 16th transformer layer and used as input. The hidden states size is a 4096 dimensional vector.

\paragraph{Grey-box baseline input}
A 13-dimensional feature vector per sentence $s_{k,i}^{\text{gen}}$ derived from MAIRA-2: token count, and mean/std/min/max of token-level logits, probabilities, and entropies.

\paragraph{PRM verifier input}
A single sequence per report $k$: clinical context $H_k$, followed by generated sentences $s_{k,i}^{\text{gen}}$ interleaved with newline separators (\verb|\n|) and (during training) ground truth labels $y_{k,i}$ ('1' or '0'). See Figure \ref{fig:prm_input_example}.

\begin{figure}[htbp]
\centering
\framebox{
  \parbox{0.9\linewidth}{
    \textbf{Prompt:}\\[0.5em] 
    Provide a description of the findings in the radiology study in comparison to the prior frontal image. INDICATION: Middle-aged man with possible pneumonia. TECHNIQUE: Anteroposterior (AP) and lateral chest radiographs. COMPARISON: Not applicable.\\[1em] 
    \textbf{Completions:}\\[0.5em] 
    "There are patchy opacities in the right upper lung, right lower lung, and left lower lung."\\[0.5em] 
    "No pleural effusion or pneumothorax."\\[0.5em] 
    "Cardiac size is normal."\\[1em] 
    \textbf{Groundtruth Labels:}\\[0.5em] 
    False, False, True
  }
}
\caption{Example PRM input structure for training, using report mimic-54422699 from the MIMIC-CXR dataset. The input includes clinical context (Indication, Technique, Comparison), MAIRA-2 generated sentences, and ground-truth labels (interleaved during training only). For inference, labels are omitted and predicted by the PRM.}
\label{fig:prm_input_example}
\end{figure}

\section{Model training details}
\label{sec:training_details_appendix}
\FloatBarrier

Trained using PyTorch \citep{Paszke2019PyTorch}, Hugging Face Transformers \citep{Wolf2019HuggingFaces} and TRL \citep{vonwerra2022trl}.

\subsection{PRM verifier}
\begin{itemize}
    \item \textbf{Base Models:} Qwen2.5-0.5B, Qwen2.5-3B \citep{Qwen2024}.
    \item \textbf{Framework:} TRL PRM Trainer.
    \item \textbf{Epochs:} 3.
    \item \textbf{Optimizer:} AdamW \citep{kingma2014adam, loshchilov2018decoupled}.
    \item \textbf{Learning Rate:} $1 \times 10^{-5}$, linear warmup/decay.
    \item \textbf{Batching:} Effective batch size 16 (2 per device, 8 accum steps).
    \item \textbf{Efficiency:} Gradient checkpointing.
    \item \textbf{Sequence Length:} Max 1024 tokens (prompt max 512).
    \item \textbf{Evaluation:} Every 50 steps on validation set.
\end{itemize}

\subsection{Baseline models}
\label{sec:baseline_training_details} 

\paragraph{ReXTrust} \citep{Hardy2024}
\begin{itemize}
    \item \textbf{Architecture:} Multi-head self-attention (proj to 1024), mean pooling, linear classifier.
    \item \textbf{Optimizer:} AdamW (wd=0.01).
    \item \textbf{Learning Rate:} $1 \times 10^{-4}$, Cosine Annealing ($T_{max}=10$) \citep{loshchilov2016sgdr}.
    \item \textbf{Batch Size:} 128.
    \item \textbf{Loss:} BCE.
    \item \textbf{Regularization:} Dropout (0.1), Early Stopping (val loss, patience=3).
\end{itemize}

\paragraph{Grey-box MLP} (Inspired by \citet{Liu2024})
\begin{itemize}
    \item \textbf{Architecture:} 2-layer MLP (13 -> 50 ReLU -> 1 logit).
    \item \textbf{Optimizer:} Adam.
    \item \textbf{Learning Rate:} Fixed $1 \times 10^{-3}$.
    \item \textbf{Epochs:} 20.
    \item \textbf{Batch Size:} 128.
    \item \textbf{Loss:} BCE.
\end{itemize}

\section{Best-of-N implementation details}
\label{sec:bon_details}
\FloatBarrier

\paragraph{Candidate generation}
$N=128$ candidates per study using MAIRA-2 (temp=1, top-p=0.9, top-k=50)

\paragraph{Scoring methods} Used for non-weighted and as base for weighted BoN (matching labels like `Avg Prob`, `Min Prob`,`Prod Prob` or `Log Prob` in Appendix \ref{app:bon_plots}):
\begin{itemize}
    \item \textbf{MinProb:} The minimum sentence probability using PRM. Score$_k = \min_{i} p_{k,i}$.
    \item \textbf{AvgProb:} The arithmetic mean of sentence probabilities using PRM. Score$_k = \frac{1}{m_k} \sum_i p_{k,i}$.
    \item \textbf{ProdProb:} The geometric mean of sentence probabilities using PRM. Score$_k = (\prod_{i=1}^{m_k} p_{k,i})^{1/m_k}$.
    \item \textbf{LogProb:} The sum of log-probabilities (equivalent to maximizing the product $\prod p_{k,i}$). Score$_k = \sum_i \log p_{k,i}$.
\end{itemize}

\paragraph{Weighted Best-of-N (PRM-based)} Adapted from \citet{li2022making}:
\begin{enumerate}
    \item \textbf{Grouping:} Assign candidates to groups $G_j$ based on identical CheXbert \citep{Smit2020} label vectors.
    \item \textbf{Intra-Group Scoring:} Calculate base score $f(S_{k,l}^{\text{gen}})$ (using MinProb, AvgProb, ProdProb, or LogProb as defined above) for each candidate $l$.
    \item \textbf{Inter-Group Aggregation:} Sum scores within each group: TotalScore$(G_j) = \sum_{l \in G_j} f(S_{k,l}^{\text{gen}})$.
    \item \textbf{Group Selection:} Select group $G_{j^*}$ with max TotalScore.
    \item \textbf{Final Selection:} Select candidate $l^*$ within $G_{j^*}$ maximizing individual score $f(S_{k,l^*}^{\text{gen}})$. (Note: For weighted plots labeled W-Avg, W-Min, W-Prod, W-Log, the base score $f$ corresponds to AvgProb, MinProb, ProdProb, LogProb respectively).
\end{enumerate}

\section{Qualitative examples of PRM verification}
\label{sec:appendix_qualitative}
\FloatBarrier

Examples of Qwen2.5-0.5B-PRM predictions on MAIRA-2 generated findings. Background: \textcolor{green!20}{Verifier Correct}, \textcolor{red!20}{Verifier Incorrect}. Border: \fcolorbox{green!50}{white}{GT Correct}, \fcolorbox{red!50}{white}{GT Incorrect}. \underline{Underlined}: Verifier prediction differs from GT. $p_{\phi, k, i}$ shown.

\begin{figure}[htbp]
\centering
\noindent 
\begin{minipage}[t]{0.28\textwidth}\vspace{0pt} 
    \includegraphics[width=\textwidth]{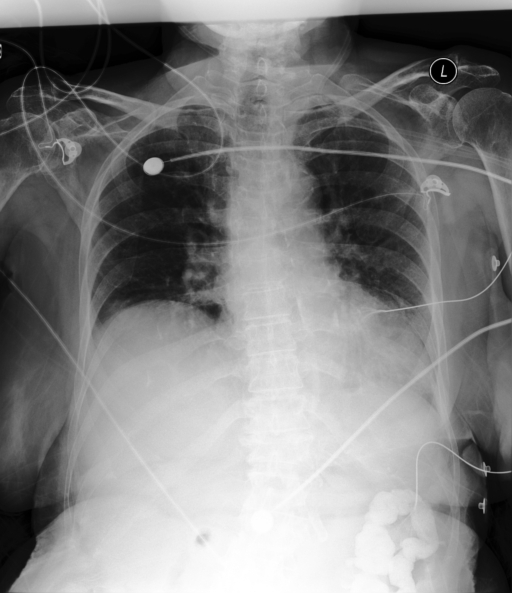}
\end{minipage}\hfill 
\begin{minipage}[t]{0.68\textwidth}\vspace{0pt} 
    \textbf{Probabilities:} 0.480, 0.786, 0.103, 0.082 
    \\
    \textbf{Report:}\\[0.5em] 
    \fcolorbox{red!50}{red!20}{\parbox{\dimexpr\linewidth-2\fboxsep}{The lungs are clear.}}\\[0.3em] 
    \fcolorbox{green!50}{green!20}{\parbox{\dimexpr\linewidth-2\fboxsep}{Negative for pleural effusion or pneumothorax.}}\\[0.3em] 
    \fcolorbox{red!50}{red!20}{\parbox{\dimexpr\linewidth-2\fboxsep}{Cardiomediastinal silhouette is within normal limits.}}\\[0.3em] 
    \fcolorbox{red!50}{green!20}{\parbox{\dimexpr\linewidth-2\fboxsep}{\underline{There are atherosclerotic aortic calcifications.}}} 
\end{minipage}
\caption{Qualitative Example 1 (mimic-56486000). Mostly correct, one false positive.}
\label{fig:qual_ex1}
\end{figure}

\begin{figure}[htbp]
\centering
\noindent
\begin{minipage}[t]{0.28\textwidth}\vspace{0pt}
    \includegraphics[width=\textwidth]{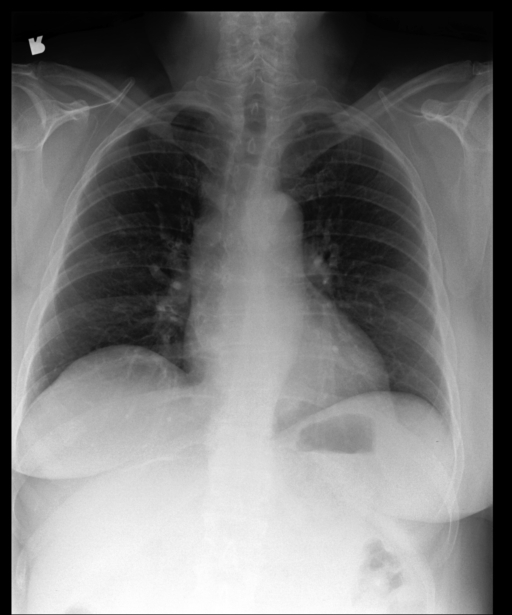}
\end{minipage}\hfill
\begin{minipage}[t]{0.68\textwidth}\vspace{0pt}
    \textbf{Probs:} 0.229, 0.913, 0.941, 0.881, 0.896, 0.525, 0.058, 0.094 
    \\
    \textbf{Report:}\\[0.5em] 
    \fcolorbox{red!50}{red!20}{\parbox{\dimexpr\linewidth-2\fboxsep}{The lungs are clear.}}\\[0.3em] 
    \fcolorbox{green!50}{green!20}{\parbox{\dimexpr\linewidth-2\fboxsep}{No focal consolidation.}}\\[0.3em] 
    \fcolorbox{green!50}{green!20}{\parbox{\dimexpr\linewidth-2\fboxsep}{No effusion.}}\\[0.3em] 
    \fcolorbox{green!50}{green!20}{\parbox{\dimexpr\linewidth-2\fboxsep}{No pneumothorax.}}\\[0.3em] 
    \fcolorbox{green!50}{green!20}{\parbox{\dimexpr\linewidth-2\fboxsep}{The cardiomediastinal silhouette is normal.}}\\[0.3em] 
    \fcolorbox{red!50}{green!20}{\parbox{\dimexpr\linewidth-2\fboxsep}{\underline{No displaced rib fractures are noted.}}}\\[0.3em] 
    \fcolorbox{red!50}{red!20}{\parbox{\dimexpr\linewidth-2\fboxsep}{There is no evidence of a compression deformity...}}\\[0.3em] 
    \fcolorbox{red!50}{red!20}{\parbox{\dimexpr\linewidth-2\fboxsep}{The thoracic spine demonstrates mild degenerative changes.}} 
\end{minipage}
\caption{Qualitative Example 2 (mimic-58865157). Mostly correct, one false positive.}
\label{fig:qual_ex2}
\end{figure}

\begin{figure}[htbp]
\centering
\noindent
\begin{minipage}[t]{0.28\textwidth}\vspace{0pt}
    \includegraphics[width=\textwidth]{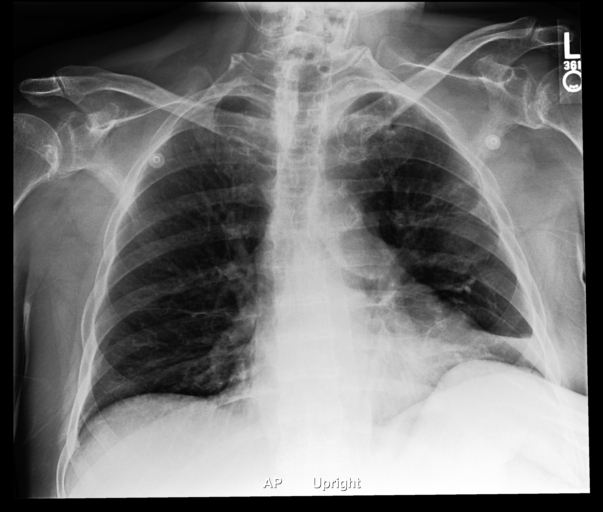}
\end{minipage}\hfill
\begin{minipage}[t]{0.68\textwidth}\vspace{0pt}
    \textbf{Probs:} 0.020, 0.955, 0.922, 0.210, 0.916 
    \\
    \textbf{Report:}\\[0.5em] 
    \fcolorbox{red!50}{red!20}{\parbox{\dimexpr\linewidth-2\fboxsep}{There is elevation of the left hemidiaphragm...}}\\[0.3em] 
    \fcolorbox{green!50}{green!20}{\parbox{\dimexpr\linewidth-2\fboxsep}{No pleural effusion.}}\\[0.3em] 
    \fcolorbox{green!50}{green!20}{\parbox{\dimexpr\linewidth-2\fboxsep}{No pneumothorax.}}\\[0.3em] 
    \fcolorbox{red!50}{red!20}{\parbox{\dimexpr\linewidth-2\fboxsep}{Normal cardiomediastinal silhouette.}}\\[0.3em] 
    \fcolorbox{green!50}{green!20}{\parbox{\dimexpr\linewidth-2\fboxsep}{There are healed left-sided rib fractures.}} 
\end{minipage}
\caption{Qualitative Example 3 (mimic-51153135). All sentences correctly verified.}
\label{fig:qual_ex3}
\end{figure}

\begin{figure}[htbp]
\centering
\noindent
\begin{minipage}[t]{0.28\textwidth}\vspace{0pt}
    \includegraphics[width=\textwidth]{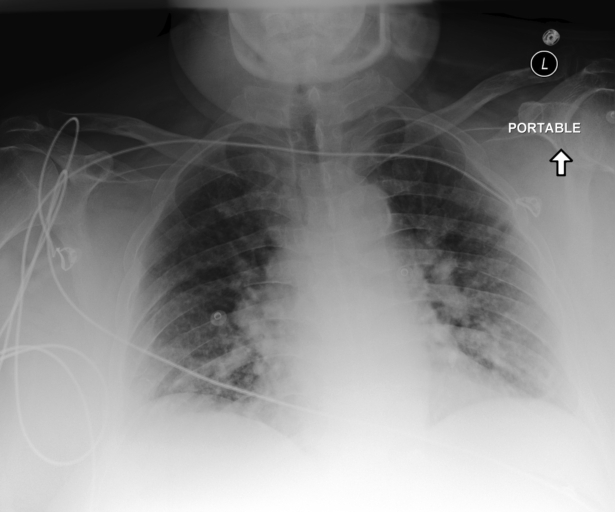}
\end{minipage}\hfill
\begin{minipage}[t]{0.68\textwidth}\vspace{0pt}
    \textbf{Probs:} 0.146, 0.049, 0.670, 0.901, 0.616 
    \\
    \textbf{Report:}\\[0.5em] 
    \fcolorbox{red!50}{red!20}{\parbox{\dimexpr\linewidth-2\fboxsep}{Mild to moderate cardiomegaly is present.}}\\[0.3em] 
    \fcolorbox{red!50}{red!20}{\parbox{\dimexpr\linewidth-2\fboxsep}{There is mild perihilar vascular congestion...}}\\[0.3em] 
    \fcolorbox{green!50}{green!20}{\parbox{\dimexpr\linewidth-2\fboxsep}{There is no focal lung consolidation.}}\\[0.3em] 
    \fcolorbox{green!50}{green!20}{\parbox{\dimexpr\linewidth-2\fboxsep}{There is no pneumothorax.}}\\[0.3em] 
    \fcolorbox{green!50}{green!20}{\parbox{\dimexpr\linewidth-2\fboxsep}{There is no large pleural effusion.}} 
\end{minipage}
\caption{Qualitative Example 4 (mimic-54259835). All sentences correctly verified.}
\label{fig:qual_ex4}
\end{figure}

\begin{figure}[htbp]
\centering
\noindent
\begin{minipage}[t]{0.28\textwidth}\vspace{0pt}
    \includegraphics[width=\textwidth]{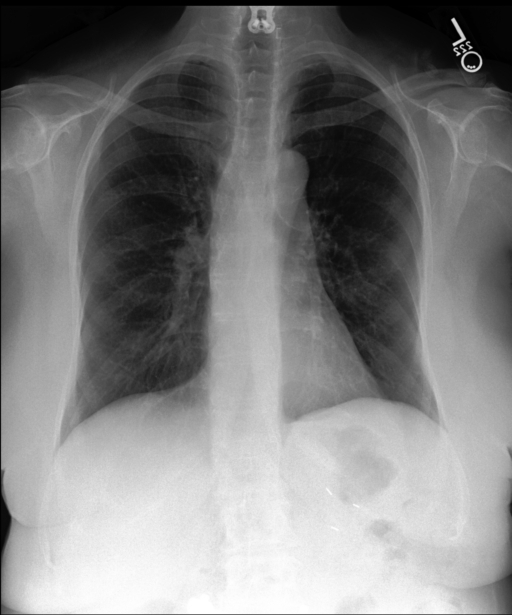}
\end{minipage}\hfill
\begin{minipage}[t]{0.68\textwidth}\vspace{0pt}
    \textbf{Probs:} 0.220, 0.582, 0.810, 0.794, 0.422, 0.140, 0.206
    \\
    \textbf{Report:}\\[0.5em] 
    \fcolorbox{green!50}{red!20}{\parbox{\dimexpr\linewidth-2\fboxsep}{\underline{The lungs are hyperinflated but clear.}}}\\[0.3em] 
    \fcolorbox{green!50}{green!20}{\parbox{\dimexpr\linewidth-2\fboxsep}{No focal consolidations.}}\\[0.3em] 
    \fcolorbox{green!50}{green!20}{\parbox{\dimexpr\linewidth-2\fboxsep}{No pneumothoraces.}}\\[0.3em] 
    \fcolorbox{red!50}{green!20}{\parbox{\dimexpr\linewidth-2\fboxsep}{\underline{No pleural effusions.}}}\\[0.3em] 
    \fcolorbox{red!50}{red!20}{\parbox{\dimexpr\linewidth-2\fboxsep}{Cardiomediastinal silhouette is within normal limits.}}\\[0.3em] 
    \fcolorbox{green!50}{red!20}{\parbox{\dimexpr\linewidth-2\fboxsep}{\underline{Mild degenerative changes of the spine.}}}\\[0.3em] 
    \fcolorbox{red!50}{red!20}{\parbox{\dimexpr\linewidth-2\fboxsep}{Surgical clips in the left upper quadrant.}} 
\end{minipage}
\caption{Qualitative Example 5 (mimic-59688743). Mixed predictions with false negatives and a false positive.}
\label{fig:qual_ex5}
\end{figure}

\FloatBarrier

\section{Additional rejection curves}
\label{app:all_metrics}
\FloatBarrier

\begin{figure*}[htbp]
  \centering
  \includegraphics[width=.32\linewidth]{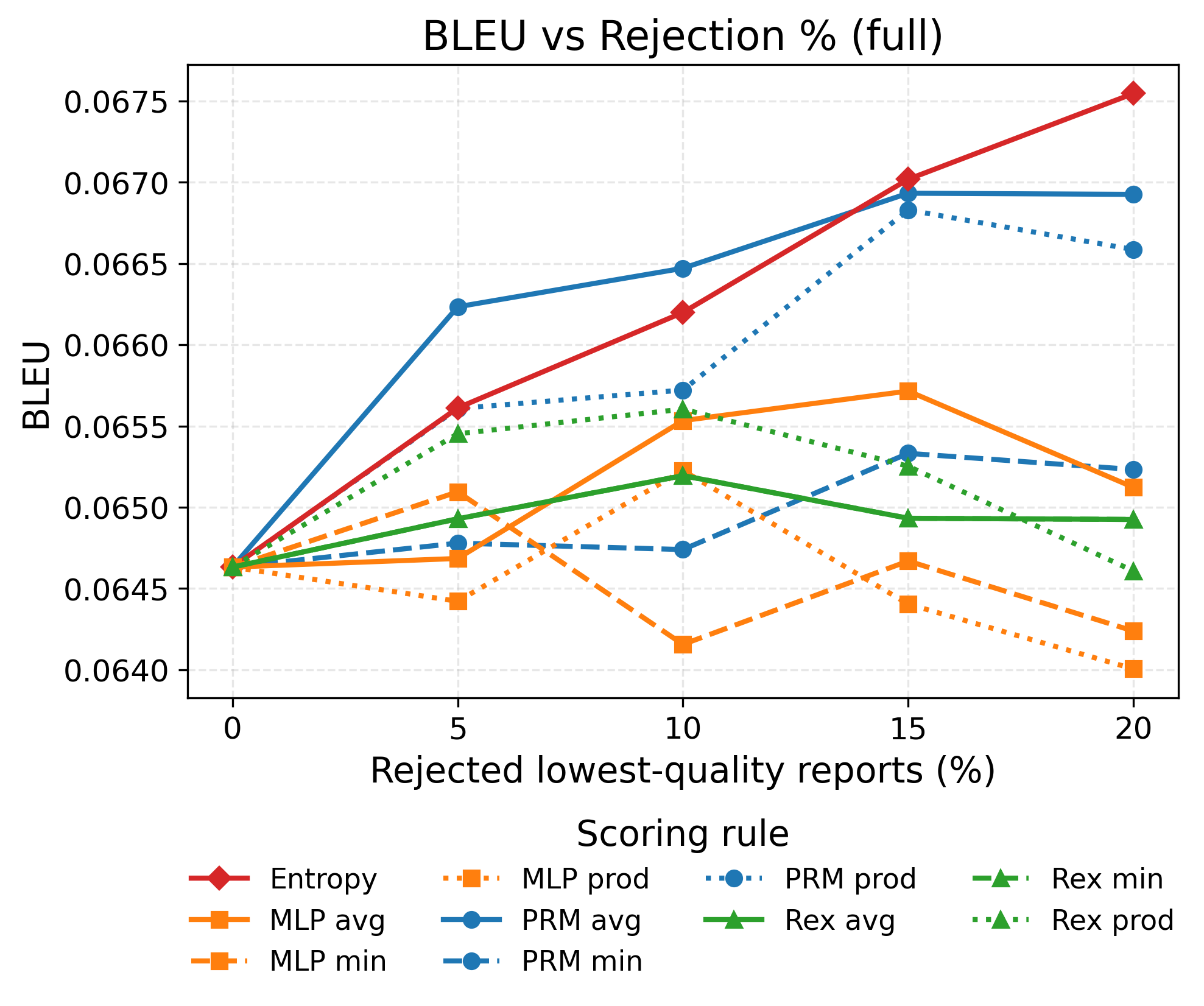}
  \includegraphics[width=.32\linewidth]{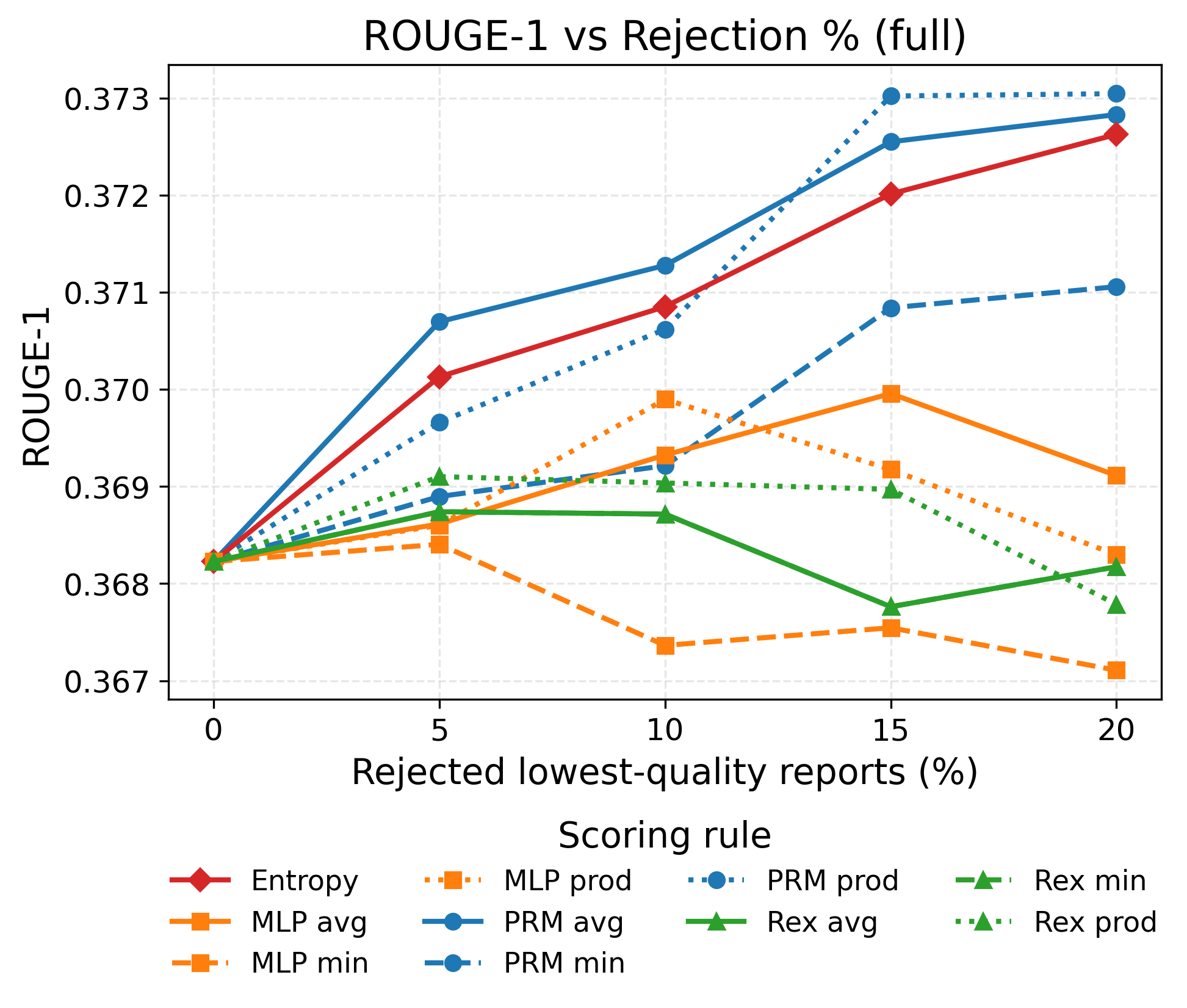}
  \includegraphics[width=.32\linewidth]{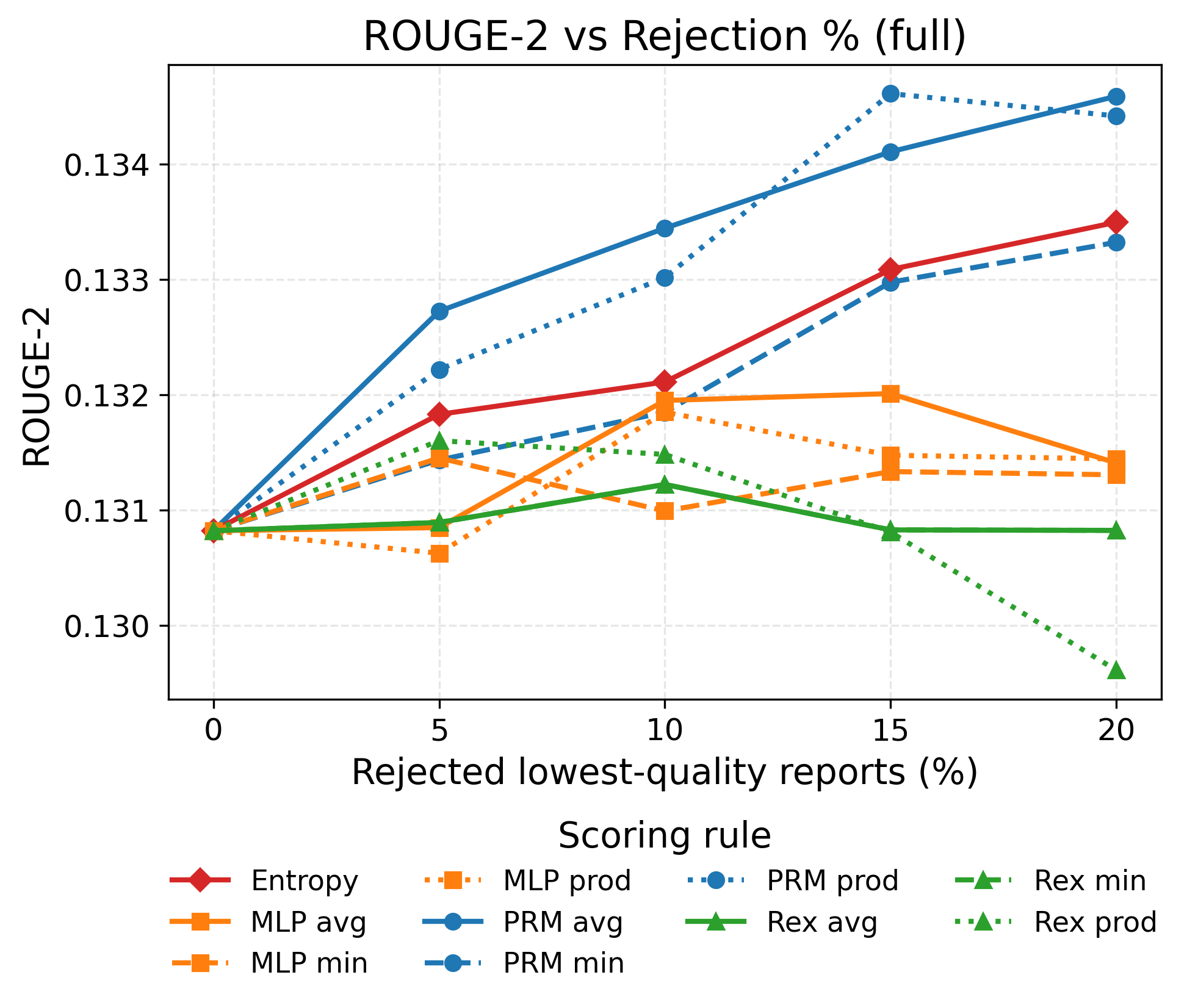}
  \includegraphics[width=.32\linewidth]{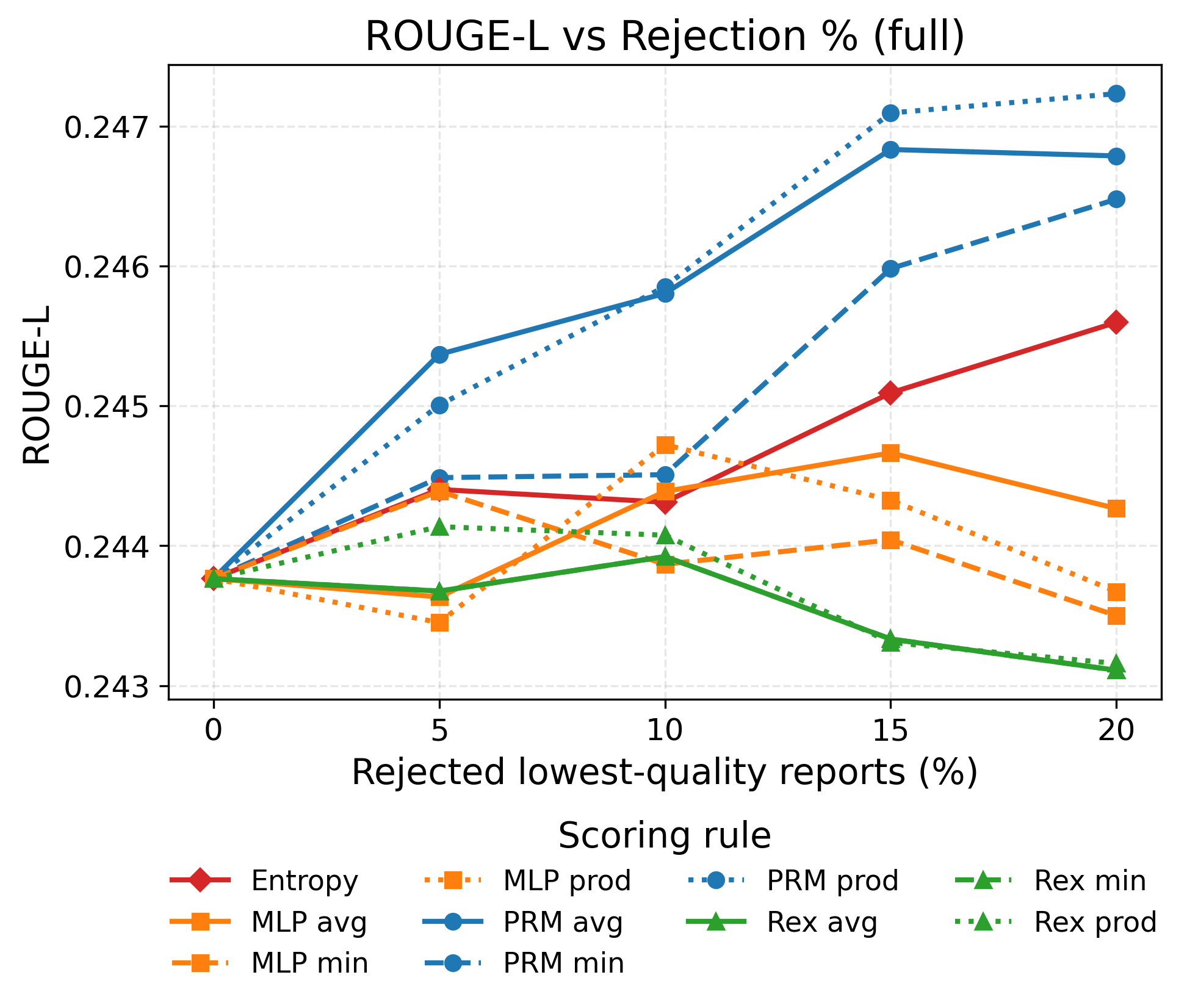}
  \includegraphics[width=.32\linewidth]{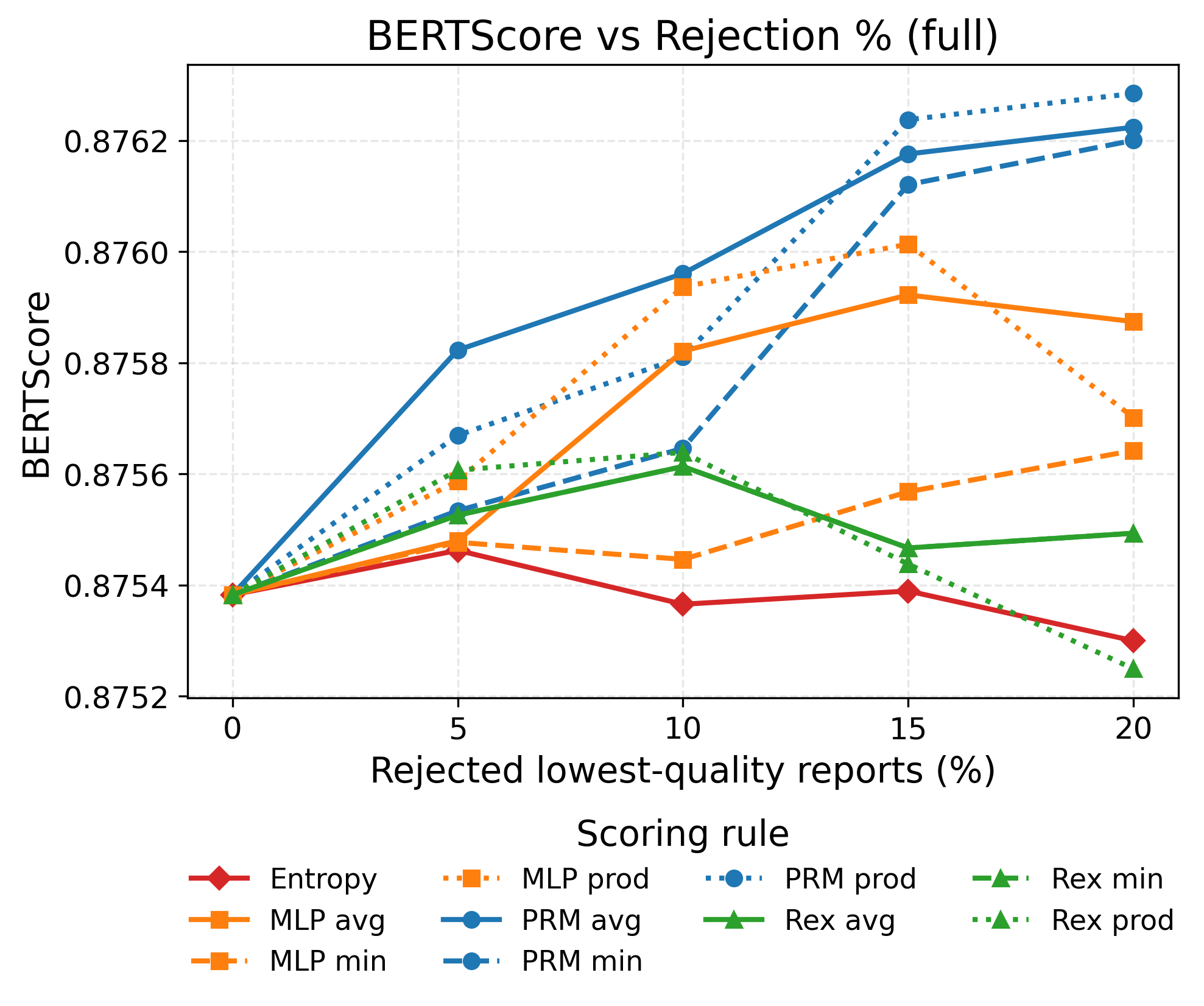}
  \includegraphics[width=.32\linewidth]{Figures/rejection/F1_RadGraph_full.png} 
  \caption{Full metric results for report rejection on the MIMIC-CXR full unbalanced test set (MAIRA-2 generated reports). The y-axis shows the metric score for the remaining reports after rejecting the x-axis percentage of lowest-scoring reports, based on various verifier aggregation rules. PRM-\textit{avg} and PRM-\textit{prod} show best improvement slopes for most metrics. Entropy as a rejection criterion also shows good improvements for BLEU and F1-RadGraph scores. (F1-CheXbert shown in main paper Fig.~\ref{fig:chex_rad_reject})}
  \label{fig:unbalanced_curves}
\end{figure*}

\begin{figure*}[htbp]
  \centering
  \includegraphics[width=.32\linewidth]{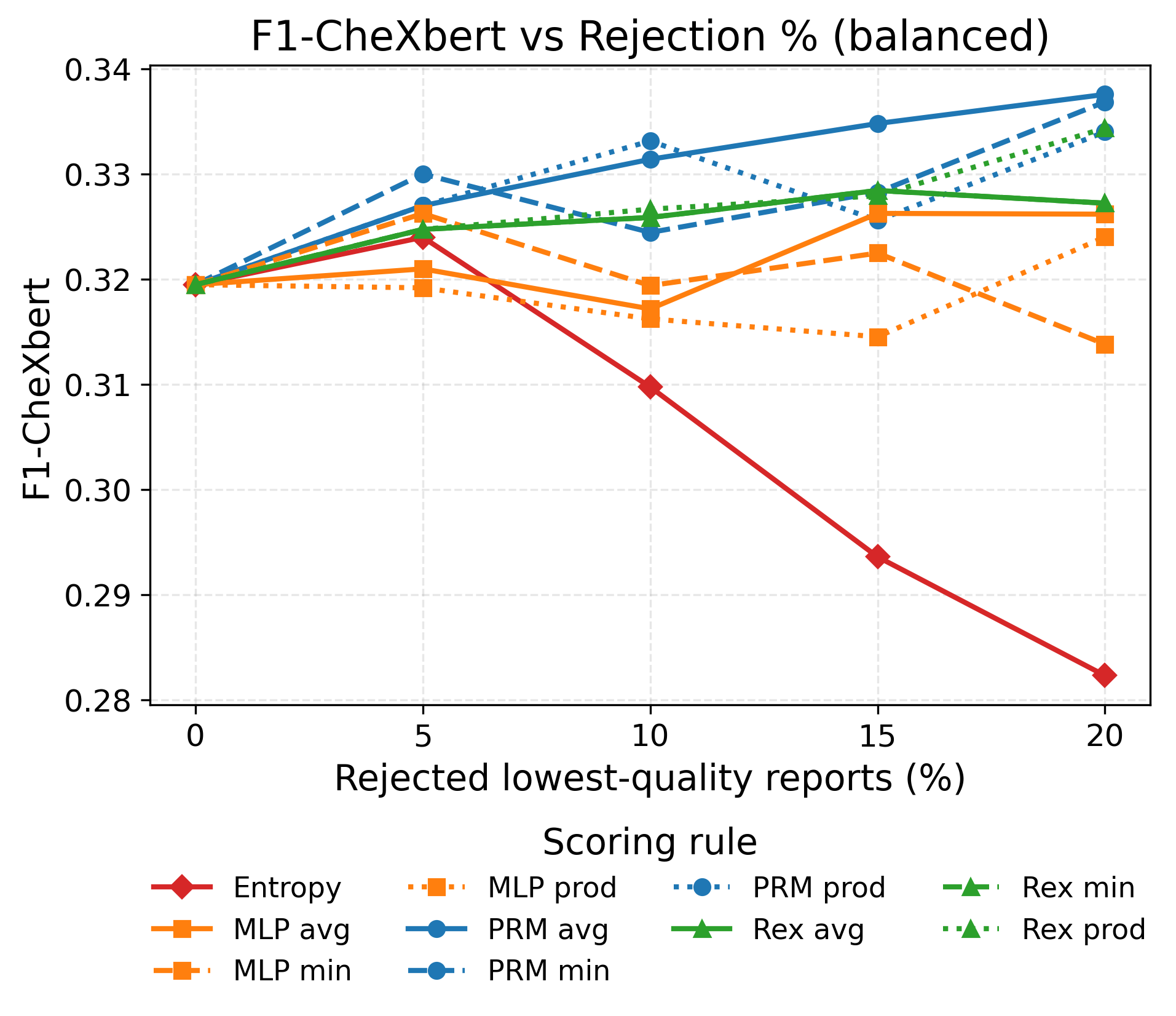}
  \includegraphics[width=.32\linewidth]{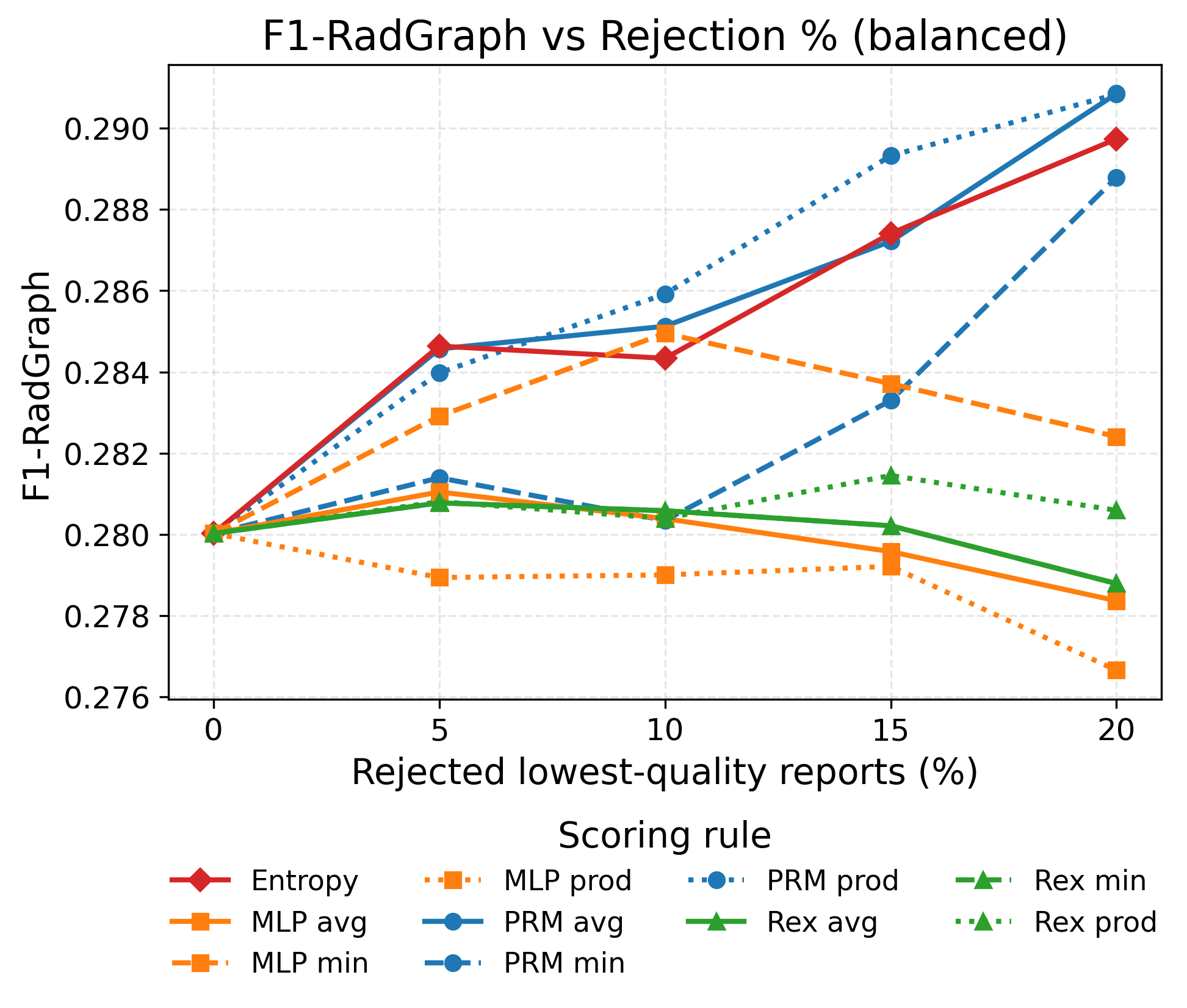}
  \includegraphics[width=.32\linewidth]{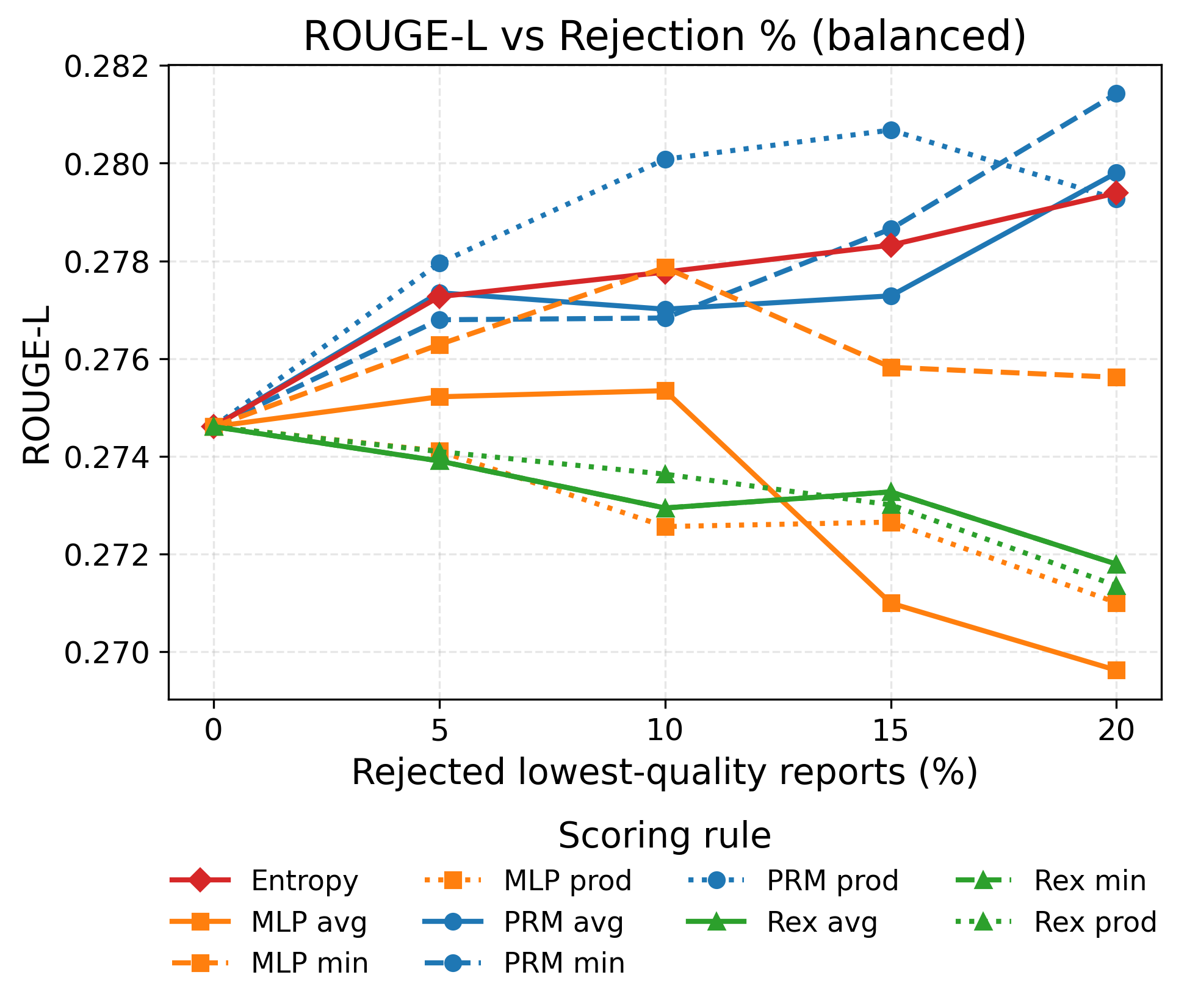}
  \caption{Selected rejection curves on the MIMIC-CXR balanced test set (MAIRA-2 generated reports). Trends mirror the unbalanced set (Figure~\ref{fig:unbalanced_curves}), confirming PRM effectiveness is robust to dataset balance.}
  \label{fig:balanced_curves}
\end{figure*}

This appendix section provides the full set of report rejection curves, complementing the \autoref{sec:results_rejection_short} of the main paper. These plots illustrate how various report-level quality metrics change as an increasing percentage of the lowest-scoring reports are discarded.
Figure \ref{fig:unbalanced_curves} shows these results on the full (unbalanced) MIMIC-CXR test set across all evaluated textual and clinical metrics.
Figure \ref{fig:balanced_curves} presents results for key metrics on the balanced MIMIC-CXR test set, demonstrating the robustness of PRM-based filtering to dataset balance.
Table \ref{tab:rejection_full_metrics_appendix} provides the precise numerical values for F1-CheXbert and F1-RadGraph scores at various rejection thresholds (0\%, 5\%, 10\%, 15\%, and 20\%) on the full unbalanced test set, complementing the visual trends shown in Figure \ref{fig:unbalanced_curves} and Figure \ref{fig:chex_rad_reject}.
The table details confirm that PRM-based scoring methods lead to notable improvements in report quality as more low-scoring reports are rejected. For instance, using PRM-\textit{avg} to reject the worst 20\% of reports increases F1-CheXbert by approximately 8.8\% (from a baseline of 0.32934 to 0.35838). For F1-RadGraph, improvements are generally more modest; PRM-\textit{prod} achieves a 1.6\% increase (from 0.22181 to 0.22541) at 20\% rejection, while Entropy scoring shows a slightly higher 1.7\% improvement (to 0.22566) at the same threshold. These results confirm the effectiveness of PRM-based methods, particularly PRM-\textit{avg} and PRM-\textit{prod}, in enhancing the clinical factuality of the reports with the verifier.

\begin{table*}[htbp]
  \centering
  \caption{Absolute F1-CheXbert and F1-RadGraph scores on the MIMIC-CXR full unbalanced test set (MAIRA-2 generated reports) after rejecting lowest-scoring reports at 0\%, 5\%, 10\%, 15\%, and 20\% percentile thresholds. Scores are shown for various aggregation methods (PRM-based, MLP-based, ReXTrust-based, and Entropy). Baseline (0\% rejection) F1-CheXbert is 0.32934 and F1-RadGraph is 0.22181.}
  \label{tab:rejection_full_metrics_appendix}
  \small 
  \resizebox{\textwidth}{!}{
  \begin{tabular}{@{}lccccccccccc@{}}
    \toprule
    \textbf{Metric} & \textbf{Rej. \%} & \textbf{PRM avg} & \textbf{PRM min} & \textbf{PRM prod} & \textbf{MLP avg} & \textbf{MLP prod} & \textbf{MLP min} & \textbf{Rex avg} & \textbf{Rex min} & \textbf{Rex prod} & \textbf{Entropy} \\
    \midrule
    F1-CheXbert & 0\%  & 0.32934 & 0.32934 & 0.32934 & 0.32934 & 0.32934 & 0.32934 & 0.32934 & 0.32934 & 0.32934 & 0.32934 \\
                & 5\%  & 0.33698 & 0.33679 & 0.33648 & 0.33270 & 0.33313 & 0.32648 & 0.32937 & 0.32937 & 0.33293 & 0.32757 \\
                & 10\% & 0.34417 & 0.33996 & 0.34242 & 0.32990 & 0.33254 & 0.32332 & 0.32543 & 0.32543 & 0.33455 & 0.32477 \\
                & 15\% & 0.35091 & 0.34008 & 0.35016 & 0.33676 & 0.33505 & 0.32588 & 0.32511 & 0.32511 & 0.32861 & 0.31992 \\
                & 20\% & 0.35838 & 0.34341 & 0.35588 & 0.33896 & 0.33732 & 0.32903 & 0.32483 & 0.32483 & 0.32635 & 0.31426 \\
    \midrule
    F1-RadGraph & 0\%  & 0.22181 & 0.22181 & 0.22181 & 0.22181 & 0.22181 & 0.22181 & 0.22181 & 0.22181 & 0.22181 & 0.22181 \\
                & 5\%  & 0.22434 & 0.22232 & 0.22307 & 0.22059 & 0.22117 & 0.22245 & 0.22169 & 0.22169 & 0.22277 & 0.22303 \\
                & 10\% & 0.22445 & 0.22258 & 0.22344 & 0.22210 & 0.22258 & 0.22244 & 0.22218 & 0.22218 & 0.22353 & 0.22316 \\
                & 15\% & 0.22499 & 0.22418 & 0.22531 & 0.22246 & 0.22272 & 0.22365 & 0.22171 & 0.22171 & 0.22178 & 0.22464 \\
                & 20\% & 0.22507 & 0.22493 & 0.22541 & 0.22209 & 0.22123 & 0.22412 & 0.22249 & 0.22249 & 0.22008 & 0.22566 \\
    \bottomrule
  \end{tabular}%
  }
  \caption*{}
\end{table*}

\FloatBarrier

\section{Additional Best-of-N results plots}
\label{app:bon_plots}
\FloatBarrier

This appendix section provides a more detailed exploration of our Best-of-N (BoN) sampling experiments, complementing Section~\ref{sec:results_bon_short}. We investigated BoN selection to assess if PRM scores could guide the selection of higher-quality reports from multiple candidates generated by MAIRA-2 (temperature 1.0, $N_{max}=128$ candidates per study from the MIMIC-CXR full unbalanced test set). Beyond standard BoN ranking using PRM scores directly (non-weighted), we explored weighted strategies (e.g., W-Avg). As detailed in Appendix~\ref{sec:bon_details}, 'W-Avg' refers to a weighted BoN approach where candidates are first grouped by their CheXbert label vectors; PRM-Avg scores are then aggregated within groups to select the best group, and finally, the top-scoring individual report from that group is chosen. This aims to leverage broader clinical finding profiles for more robust selection.

\begin{figure*}[htbp]
  \centering
  \begin{subfigure}{0.32\linewidth}\includegraphics[width=\linewidth]{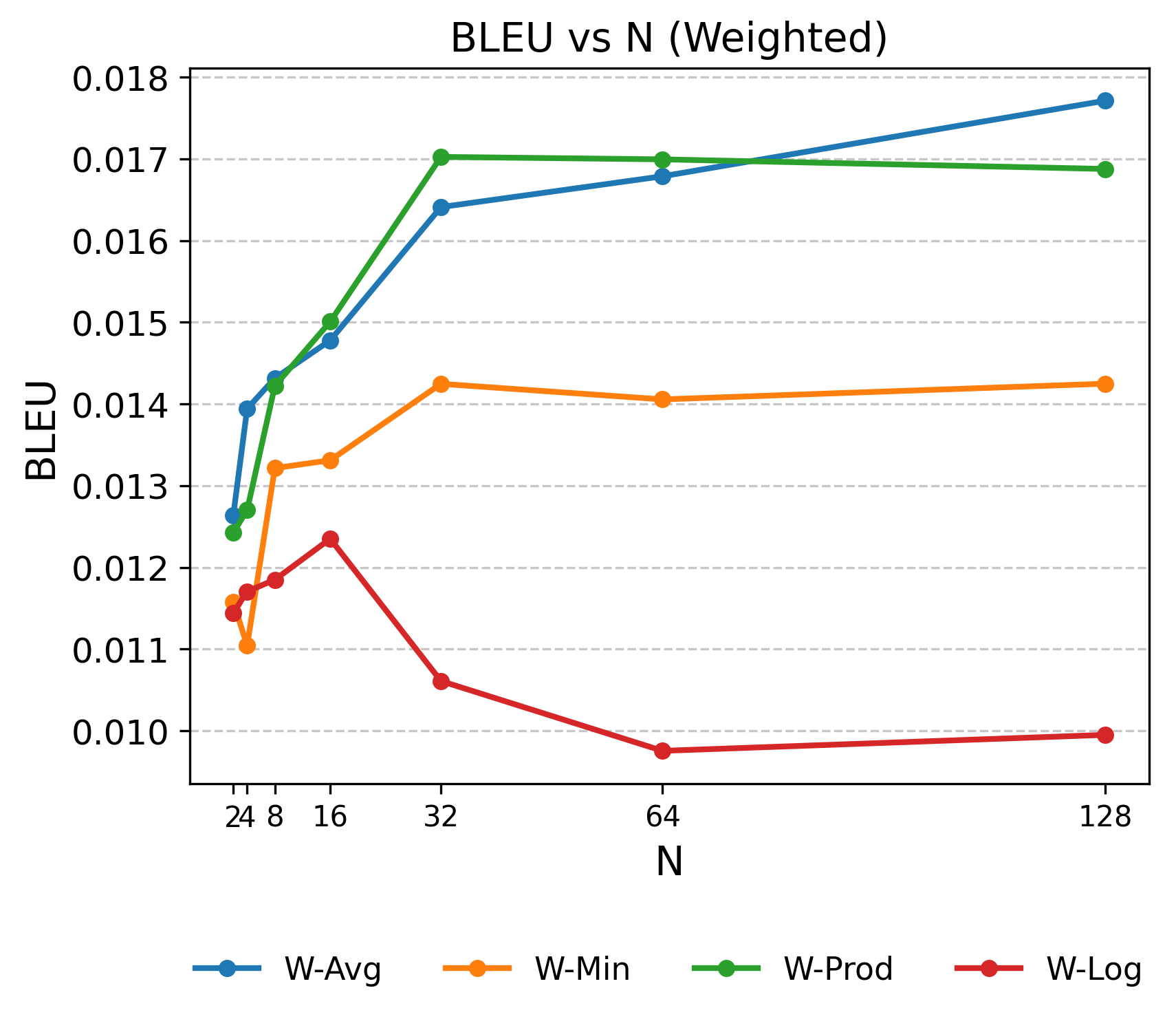}\caption{BLEU (W)}\end{subfigure}\hfill
  \begin{subfigure}{0.32\linewidth}\includegraphics[width=\linewidth]{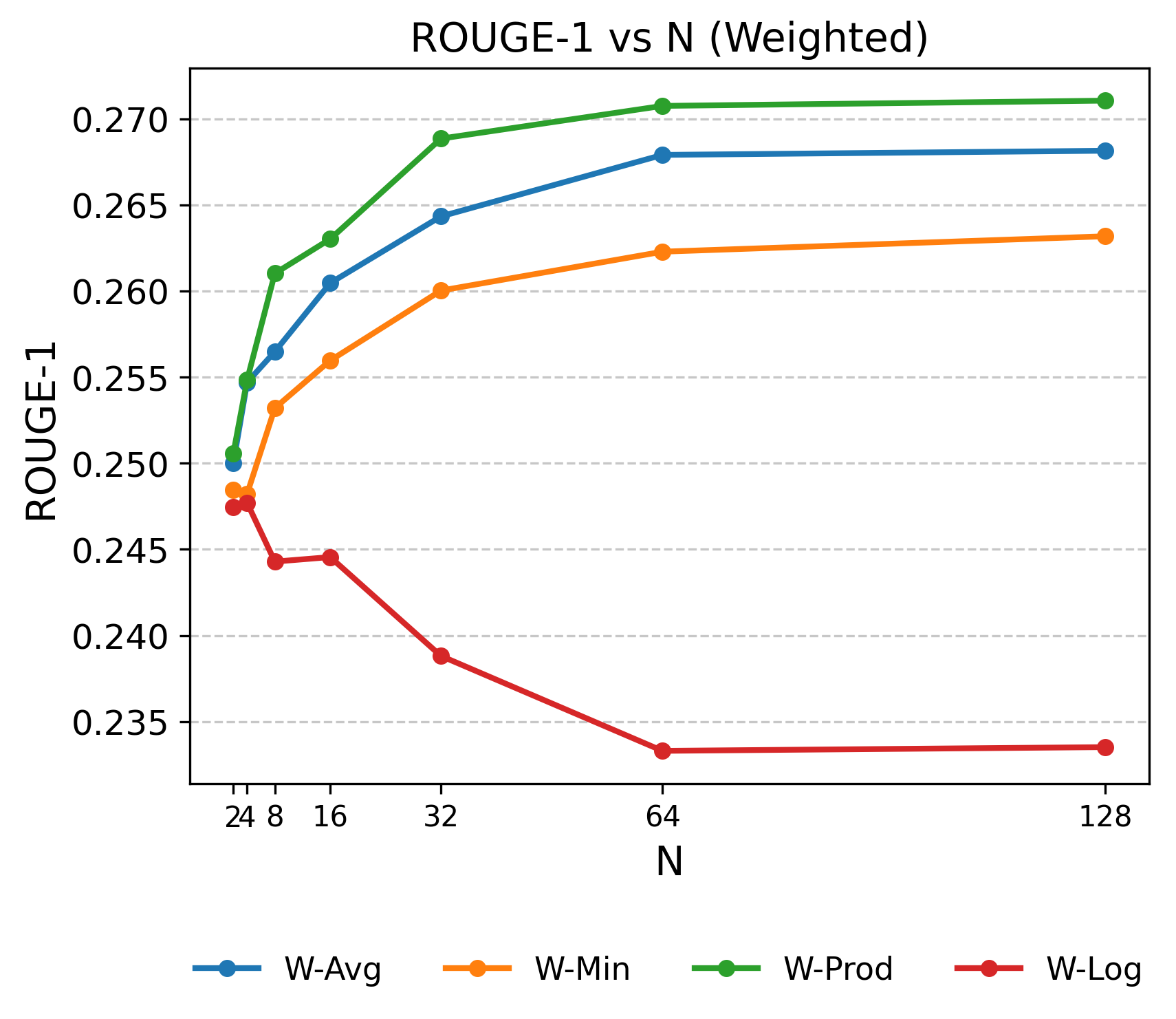}\caption{ROUGE-1 (W)}\end{subfigure}\hfill
  \begin{subfigure}{0.32\linewidth}\includegraphics[width=\linewidth]{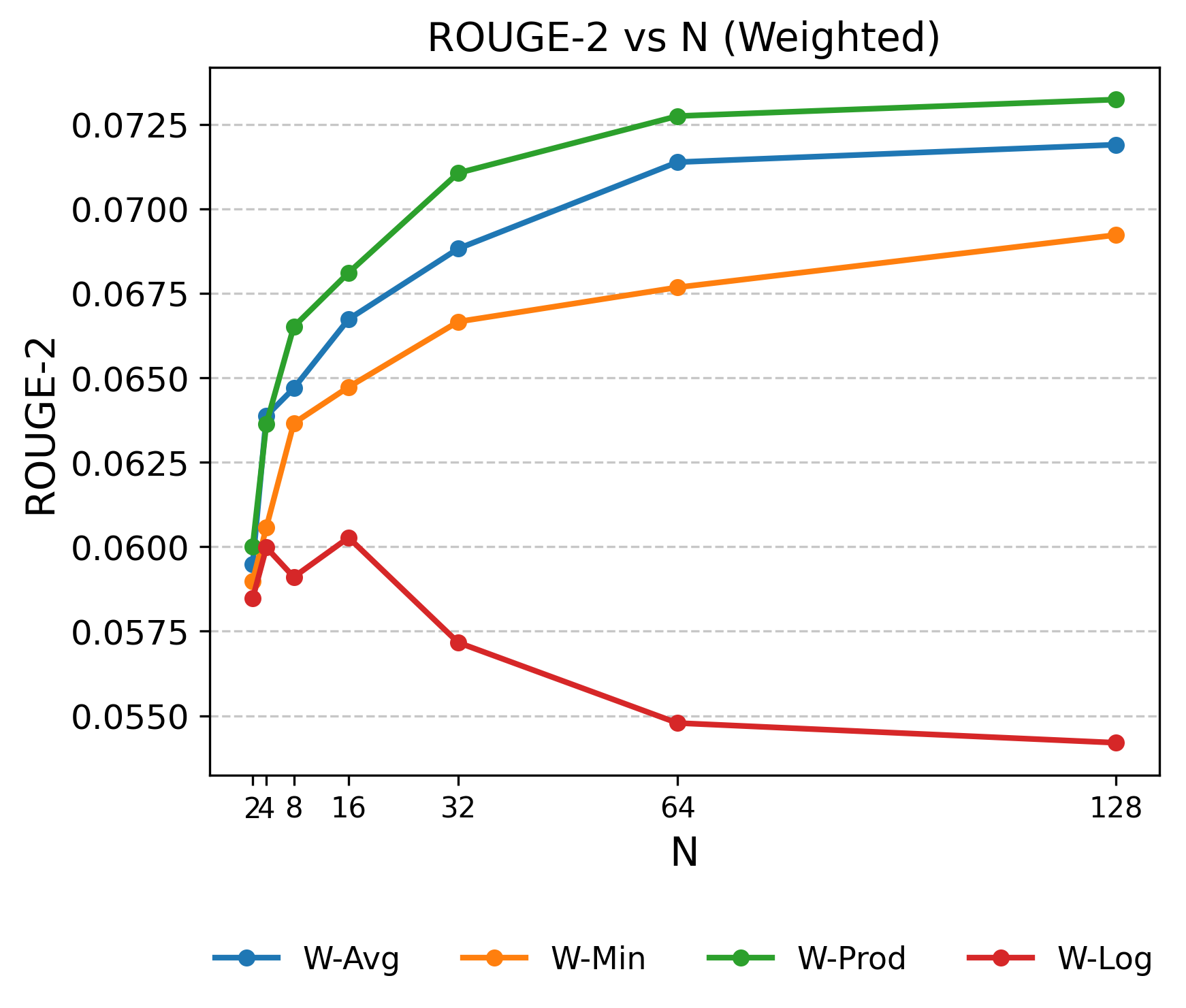}\caption{ROUGE-2 (W)}\end{subfigure}
  \begin{subfigure}{0.32\linewidth}\includegraphics[width=\linewidth]{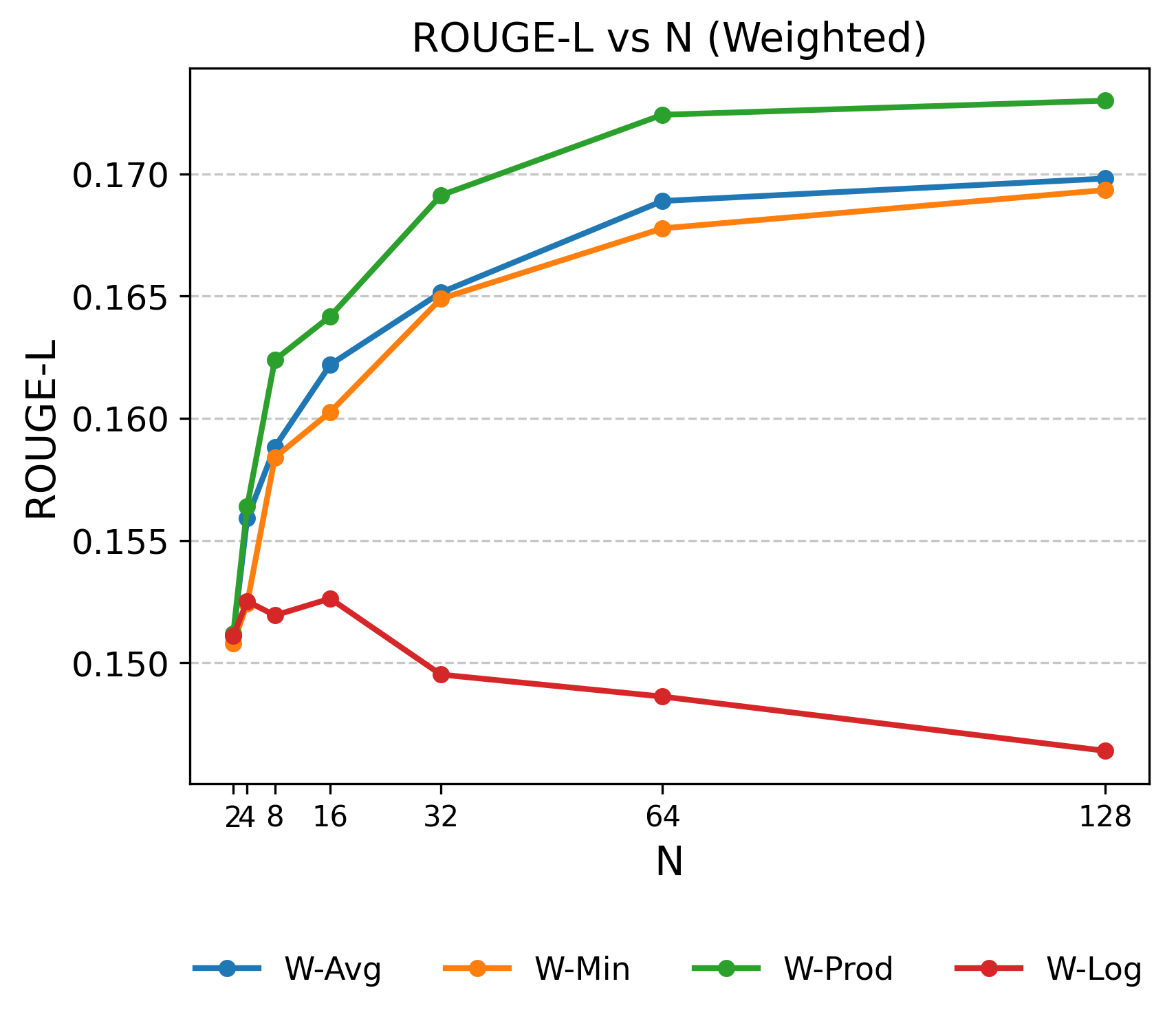}\caption{ROUGE-L (W)}\end{subfigure}\hfill
  \begin{subfigure}{0.32\linewidth}\includegraphics[width=\linewidth]{Figures/bestN/w_bertscore.png}\caption{BERTScore (W)}\end{subfigure}\hfill 
  \begin{subfigure}{0.32\linewidth}\includegraphics[width=\linewidth]{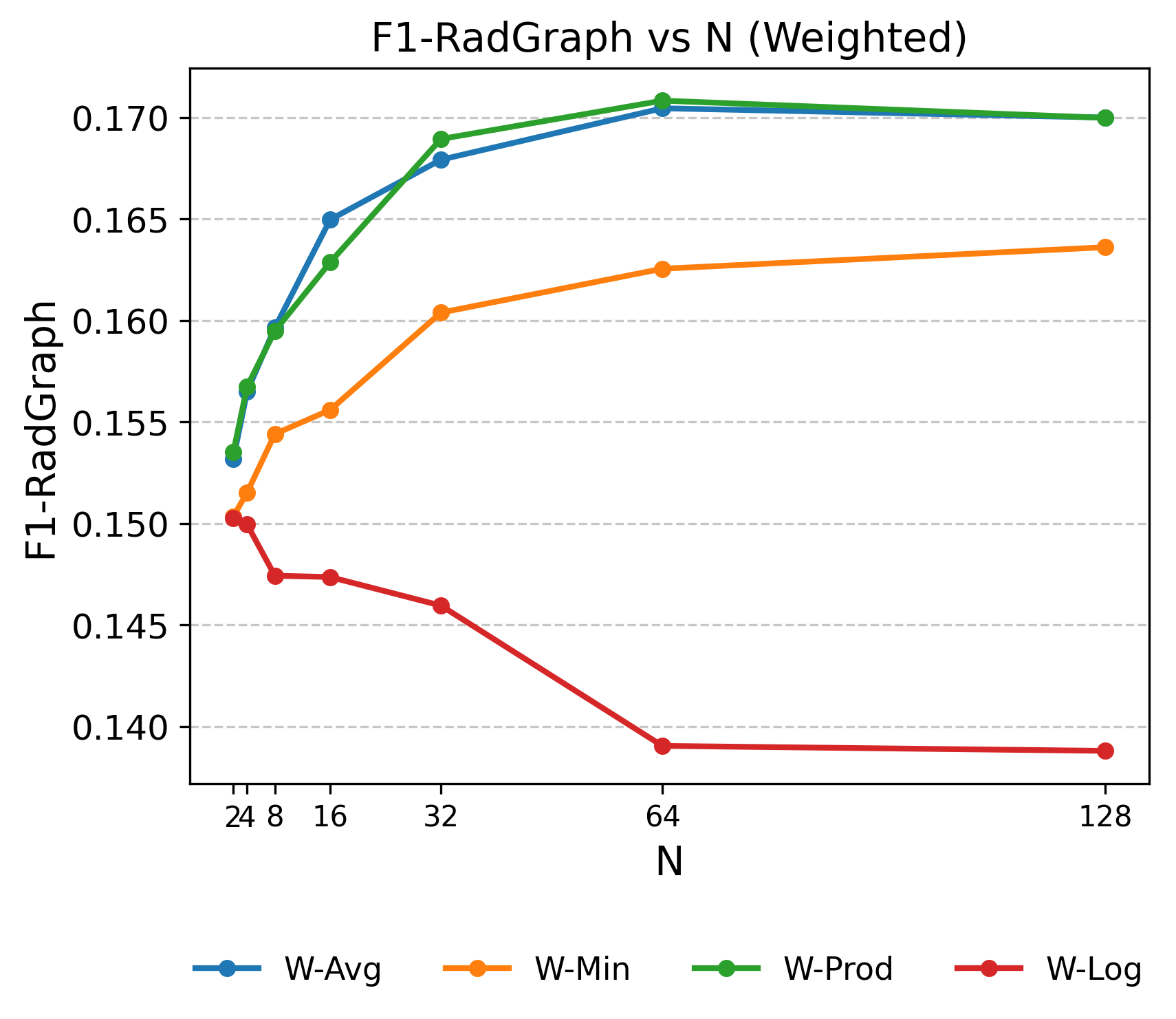}\caption{F1-RadGraph (W)}\end{subfigure}
  \begin{subfigure}{0.32\linewidth}\includegraphics[width=\linewidth]{Figures/bestN/w_chexbert.png}\caption{F1-CheXbert (W)}\end{subfigure} 
  \caption{Performance of Weighted Best-of-$N$ (BoN) selection strategies across all metrics versus $N$. Candidates were generated by MAIRA-2 (temperature 1.0) for each study in the MIMIC-CXR full unbalanced test set. Scoring strategies W-Avg, W-Min, W-Prod, and W-Log refer to weighted BoN approaches using base PRM scores from AvgProb, MinProb, ProdProb, and LogProb, respectively. (F1-CheXbert and BERTScore also shown in main paper Fig.~\ref{fig:bon_weighted_key})}
  \label{fig:bon_weighted_all}
\end{figure*}

\begin{figure*}[htbp]
  \centering
   \begin{subfigure}{0.32\linewidth}\includegraphics[width=\linewidth]{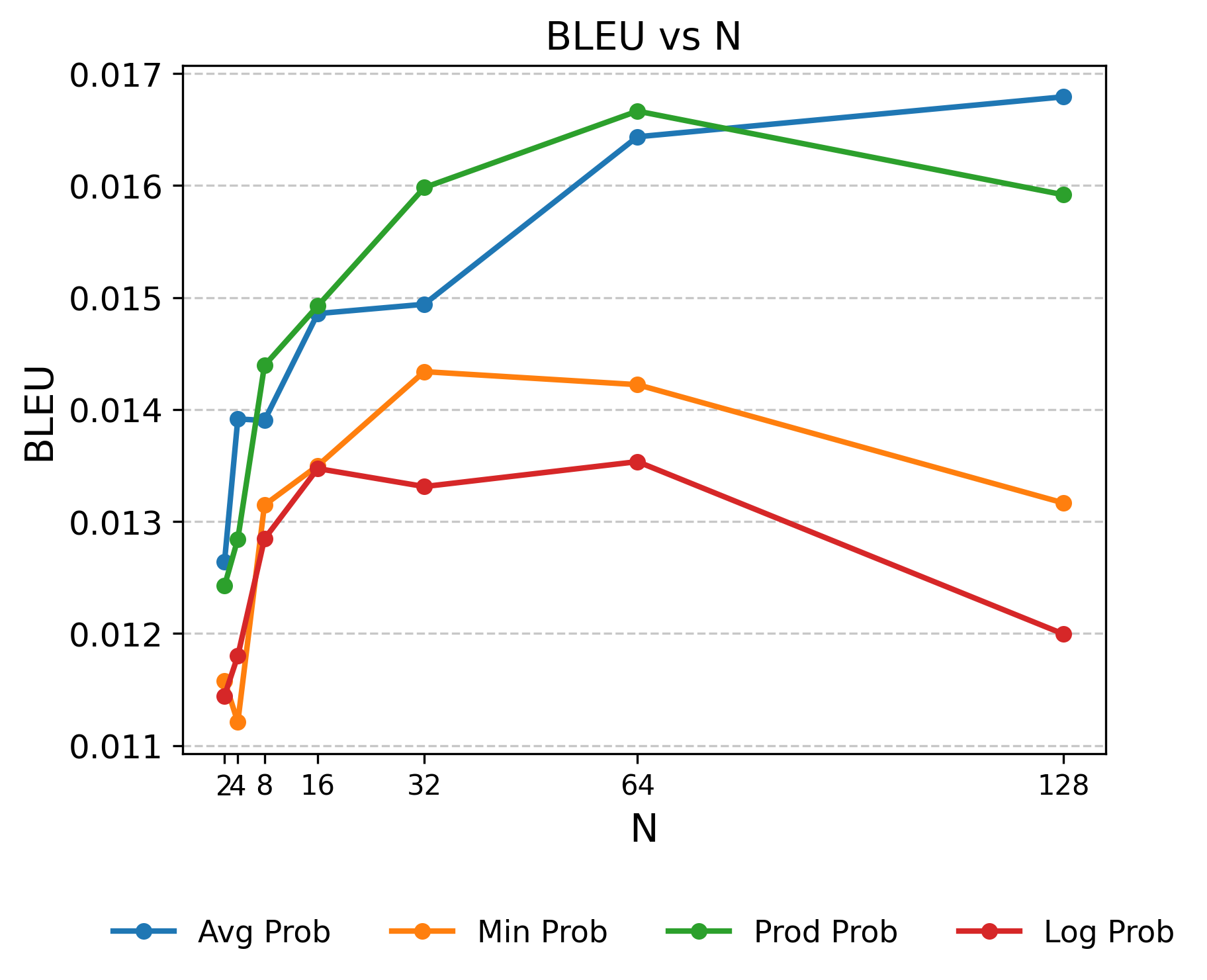}\caption{BLEU (NW)}\end{subfigure}\hfill
   \begin{subfigure}{0.32\linewidth}\includegraphics[width=\linewidth]{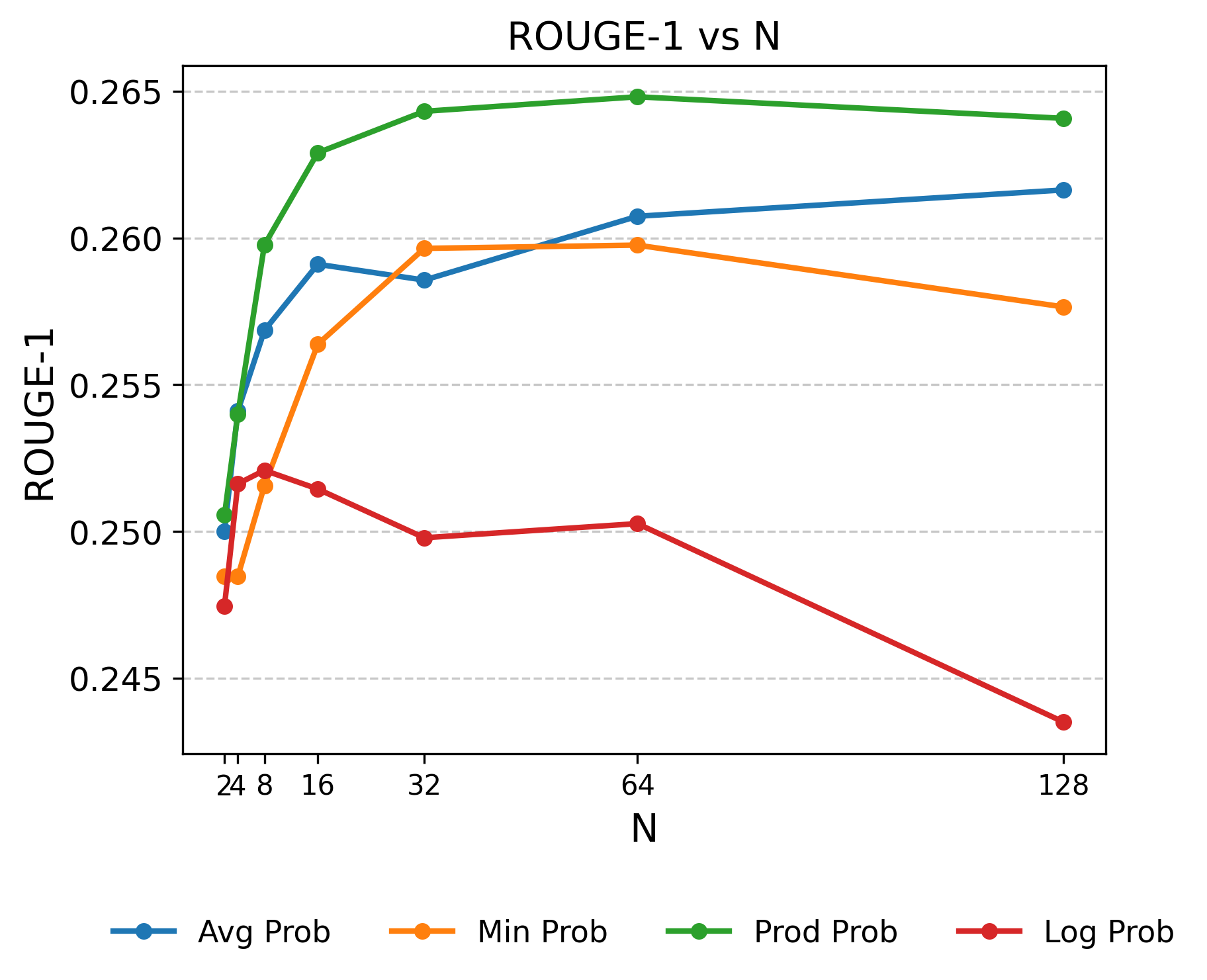}\caption{ROUGE-1 (NW)}\end{subfigure}\hfill
   \begin{subfigure}{0.32\linewidth}\includegraphics[width=\linewidth]{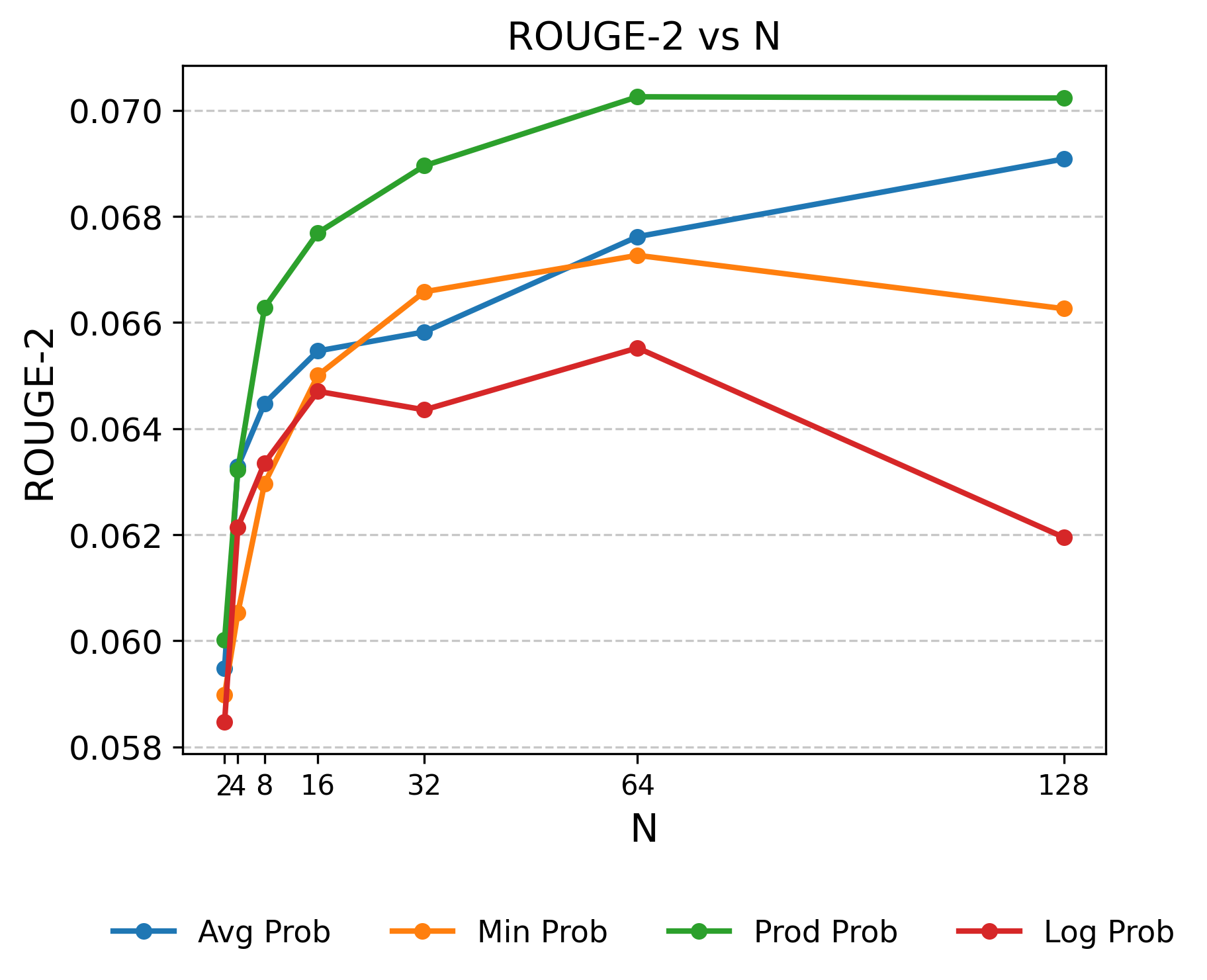}\caption{ROUGE-2 (NW)}\end{subfigure}
   \begin{subfigure}{0.32\linewidth}\includegraphics[width=\linewidth]{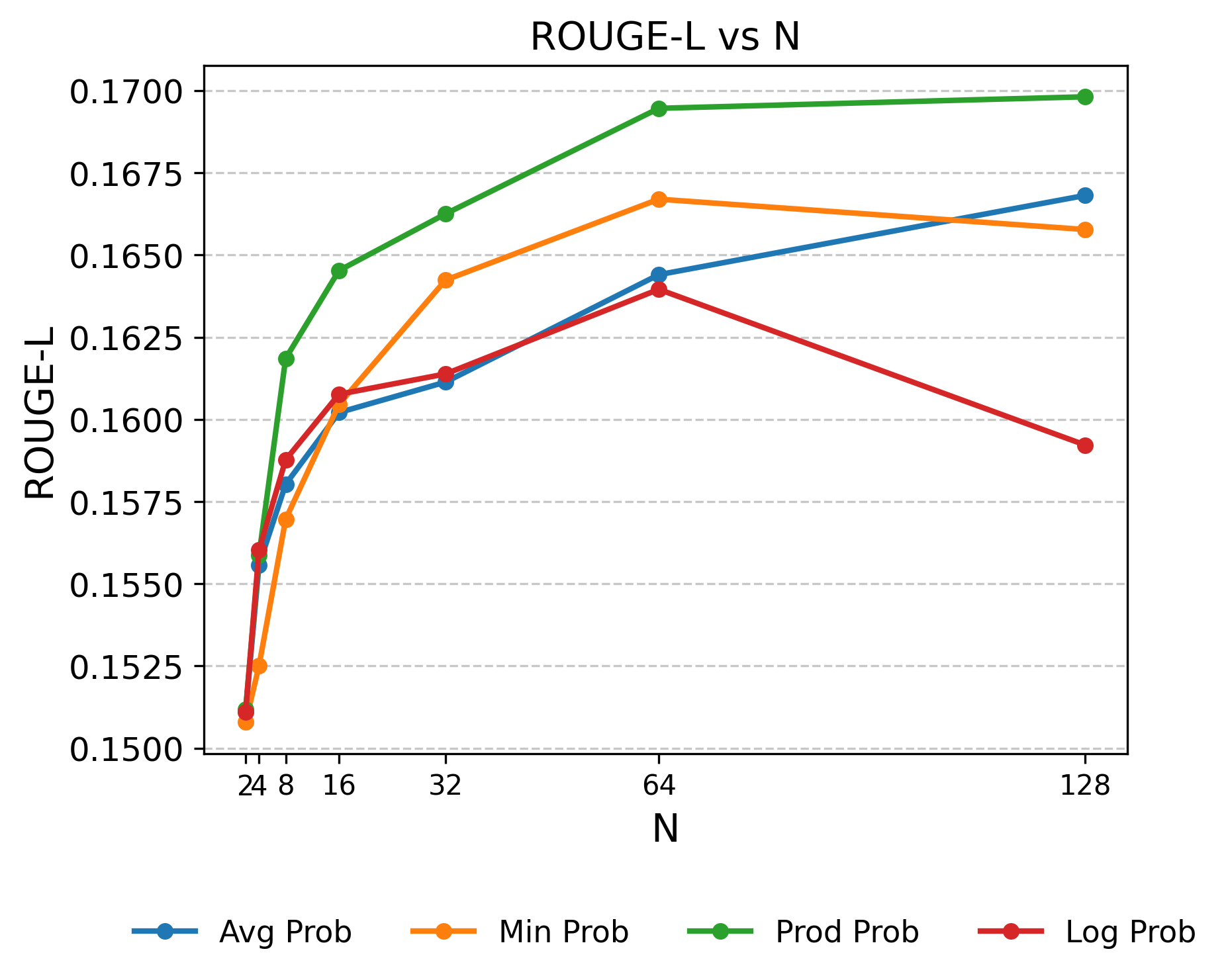}\caption{ROUGE-L (NW)}\end{subfigure}\hfill
   \begin{subfigure}{0.32\linewidth}\includegraphics[width=\linewidth]{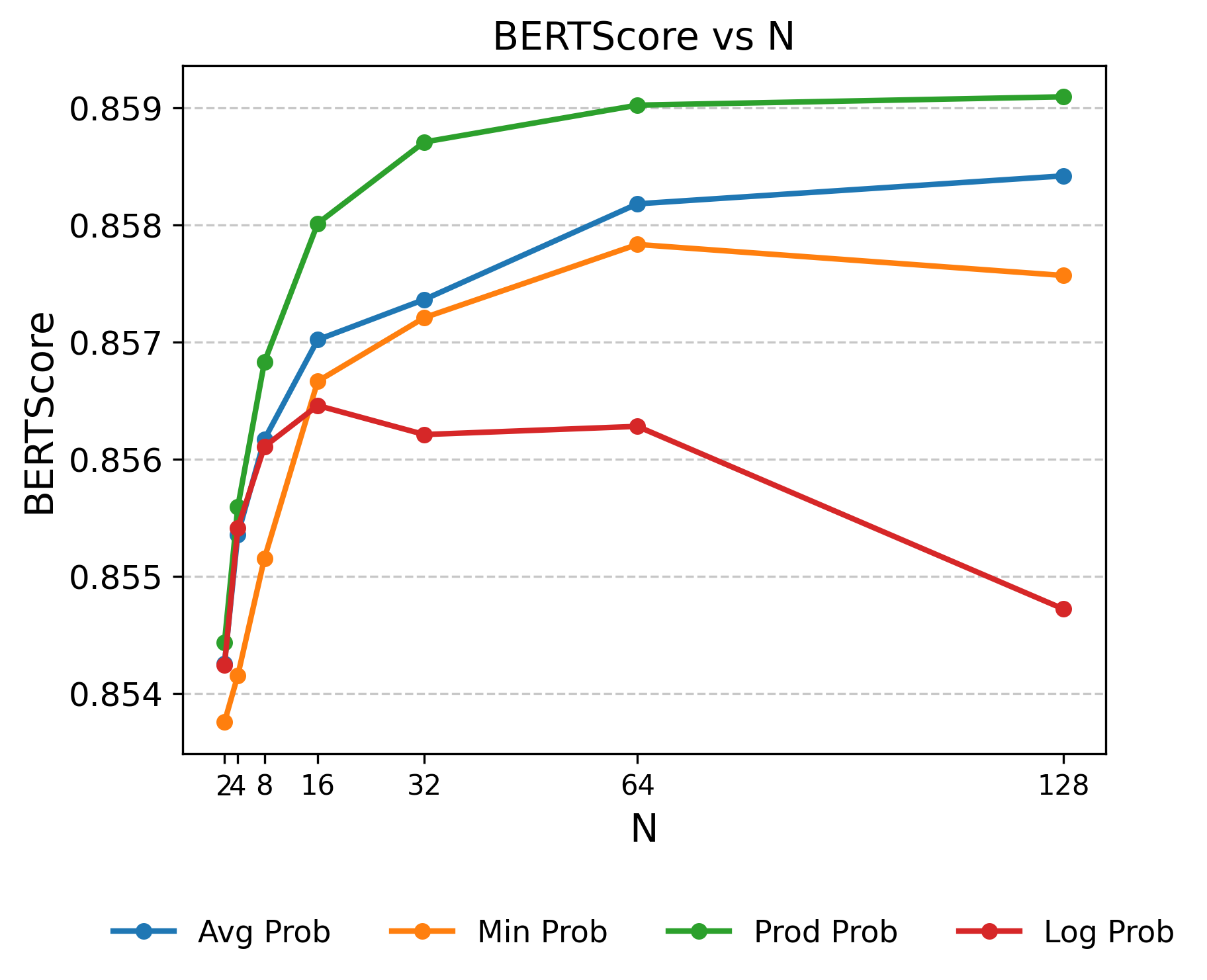}\caption{BERTScore (NW)}\end{subfigure}\hfill
   \begin{subfigure}{0.32\linewidth}\includegraphics[width=\linewidth]{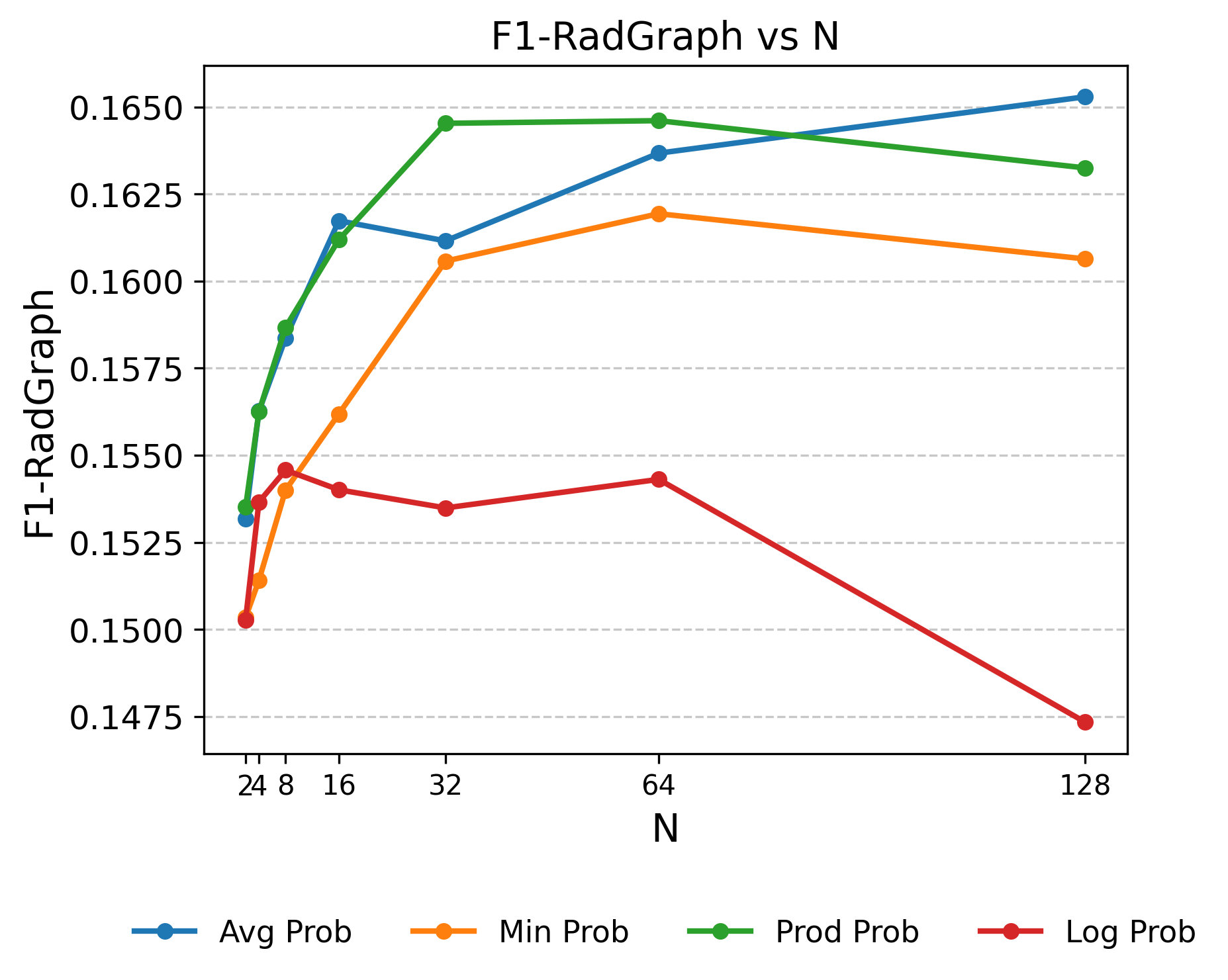}\caption{F1-RadGraph (NW)}\end{subfigure}
   \begin{subfigure}{0.32\linewidth}\includegraphics[width=\linewidth]{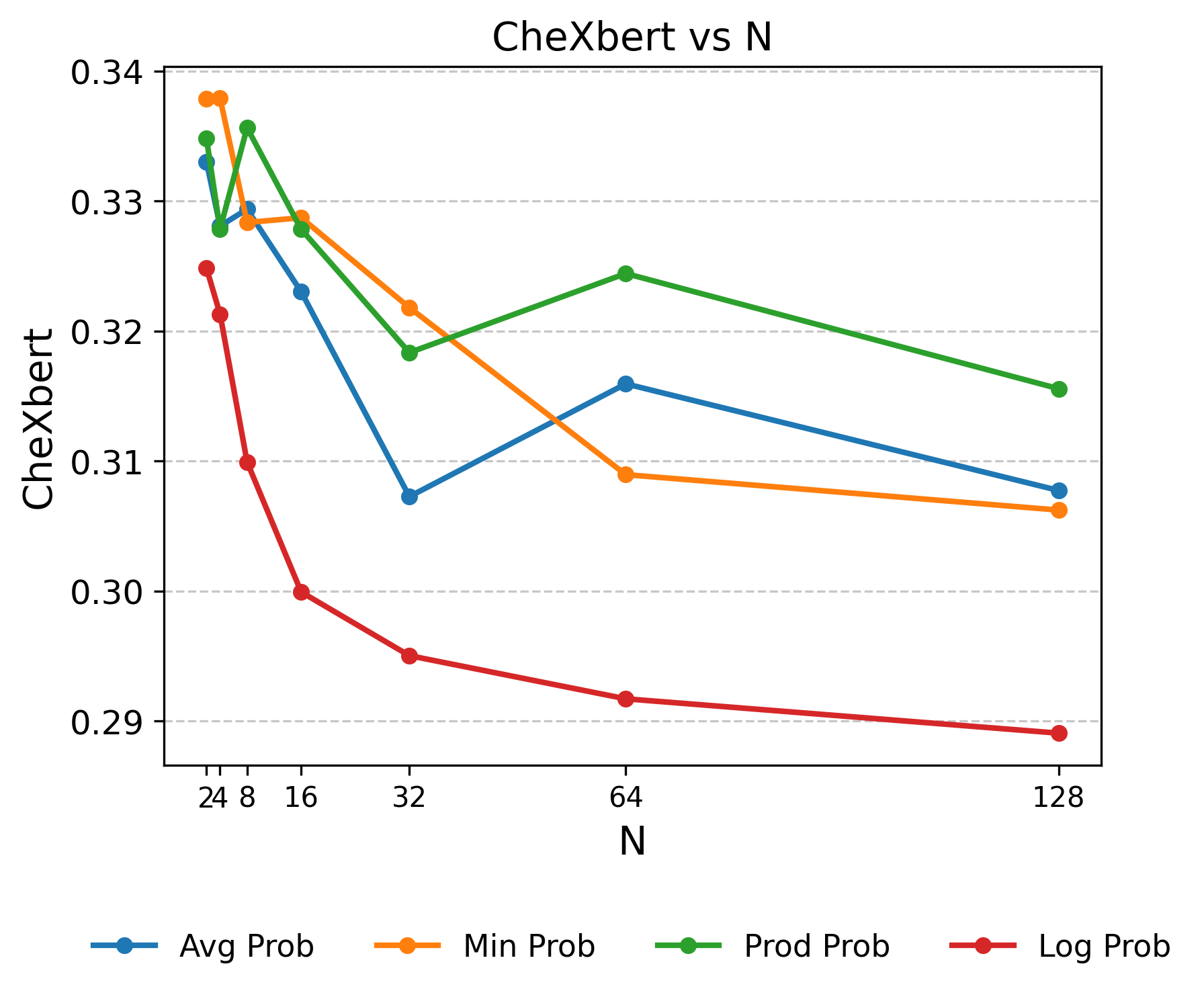}\caption{F1-CheXbert (NW)}\end{subfigure}
   \caption{Performance of Non-Weighted (NW) Best-of-$N$ (BoN) selection strategies across all metrics versus $N$. Candidates were generated by MAIRA-2 (temperature 1.0) for each study in the MIMIC-CXR full unbalanced test set. Reports are ranked directly by PRM-derived scores (AvgProb, MinProb, ProdProb) or generator LogProb.}
   \label{fig:bon_nonweighted_all}
\end{figure*}

Our results (Figures~\ref{fig:bon_weighted_all} and \ref{fig:bon_nonweighted_all}) show that weighted strategies generally outperform non-weighted ones. For instance, Weighted Avg (W-Avg) using PRM scores improves F1-CheXbert by +5.0 points (0.3575 vs 0.3077 for non-weighted AvgProb) at $N=128$. Similarly, W-Avg boosts BERTScore by +0.001 points (0.8594 vs 0.8584) under the same conditions. The choice of $N$ also impacts performance, with larger $N$ generally leads to better results.

\FloatBarrier

\end{document}